\documentclass[10pt,twocolumn,letterpaper]{article}

\usepackage[pagenumbers]{cvpr}              

\usepackage{xspace}

\usepackage{fontawesome}

\definecolor{cvprblue}{rgb}{0.21,0.49,0.74}
\usepackage[pagebackref,breaklinks,colorlinks,allcolors=cvprblue]{hyperref}
\usepackage{multicol}
\usepackage[table]{xcolor} 
\usepackage{makecell}
\definecolor{deepred}{RGB}{180,0,0}
\usepackage{wrapfig}
\usepackage[utf8]{inputenc} 
\usepackage[T1]{fontenc}    
\usepackage{hyperref}       
\usepackage{url}            
\usepackage{booktabs}       
\usepackage{arydshln}
\usepackage{amsfonts}       
\usepackage{graphicx}       
\usepackage{nicefrac}       
\usepackage{microtype}      
\usepackage{enumitem}
\usepackage{algorithm}
\usepackage{amsmath}
\usepackage{multirow}
\usepackage{pifont}
\usepackage{algorithm}
\usepackage{algpseudocode} 
\usepackage{longtable}
\usepackage{array}
\usepackage[most]{tcolorbox}
\usepackage{xcolor}

\usepackage{colortbl}   
\usepackage{arydshln}   
\usepackage{makecell}   
\usepackage{multirow}

\definecolor{light-light-gray}{gray}{0.92}

\definecolor{lightred}{RGB}{150,30,30}
\definecolor{lightblue}{RGB}{20,60,120}
\definecolor{backblue}{RGB}{236,244,252}
\definecolor{citecolor}{HTML}{0051f4}
\definecolor{pink}{HTML}{ed008c}

\hypersetup{
    colorlinks,
    linkcolor=cvprblue,
    urlcolor=cvprblue,
    citecolor=cvprblue
}

\def\paperID{*****} 
\def\confName{CVPR}
\def\confYear{2026}

\title{V-Thinker: Interactive Thinking with Images}

\author{
\textbf{Runqi Qiao}$^{1}$\thanks{Equal contribution.}~\,\thanks{Work done as interns at WeChat, Tencent Inc.}~,
\textbf{Qiuna Tan}$^{1}$\footnotemark[1]~\,\footnotemark[2]~,
\textbf{Minghan Yang}$^1$,
\textbf{Guanting Dong}$^1$,
\textbf{Peiqing Yang}$^1$,\\
\textbf{Shiqiang Lang}$^1$,
\textbf{Enhui Wan}$^1$,
\textbf{Xiaowan Wang}$^1$,
\textbf{Yida Xu}$^1$,\\
\textbf{Lan Yang}$^1$,
\textbf{Chong Sun}$^2$,
\textbf{Chen Li}$^2$\thanks{Corresponding author. zhhg@bupt.edu.cn; chaselli@tencent.com}~,
\textbf{Jing LYU}$^2$,
\textbf{Honggang Zhang}$^1$\footnotemark[3]
\\
$^1$Beijing University of Posts and Telecommunications \\
$^2$WeChat Vision, Tencent Inc. \\
\texttt{qrq@bupt.edu.cn} \quad \texttt{qiunatan@bupt.edu.cn}\\
\textbf{\url{https://github.com/We-Math/V-Thinker}} 
}

\begin{document}

\maketitle

\begin{abstract}
Empowering Large Multimodal Models (LMMs) to deeply integrate image interaction with long-horizon reasoning capabilities remains a long-standing witness in this field. Recent advances in vision-centric reasoning explore a promising ``Thinking with Images'' paradigm for LMMs, profoundly shifting from image-assisted reasoning to image-interactive thinking. While this milestone enables models to focus on fine-grained image regions, progress remains constrained by narrow visual tool spaces and task-specific workflow designs. To bridge this gap, we present \textbf{V-Thinker}, a general-purpose multimodal reasoning assistant that enables interactive, vision-centric thinking through end-to-end reinforcement learning. V-Thinker comprises two key components: (1) a \textbf{Data Evolution Flywheel} that automatically synthesizes, evolves, and verifies interactive reasoning datasets across three dimensions—diversity, quality, and difficulty; and (2) a \textbf{Visual Progressive Training Curriculum} that first aligns perception via point-level supervision, then integrates interactive reasoning through a two-stage reinforcement learning framework. 
Furthermore, we introduce \textbf{VTBench}, an expert-verified benchmark targeting vision-centric interactive reasoning tasks. Extensive experiments demonstrate that V-Thinker consistently outperforms strong LMM-based baselines in both general and interactive reasoning scenarios, providing valuable insights for advancing image-interactive reasoning applications.
\end{abstract}

\section{Introduction}
\label{intro}

\begin{quote}
\textit{"The soul never thinks without an image." \\  --- Aristotle}
\end{quote}

\begin{figure}[t!]
    \centering
    \resizebox{0.46\textwidth}{!}{
    \includegraphics{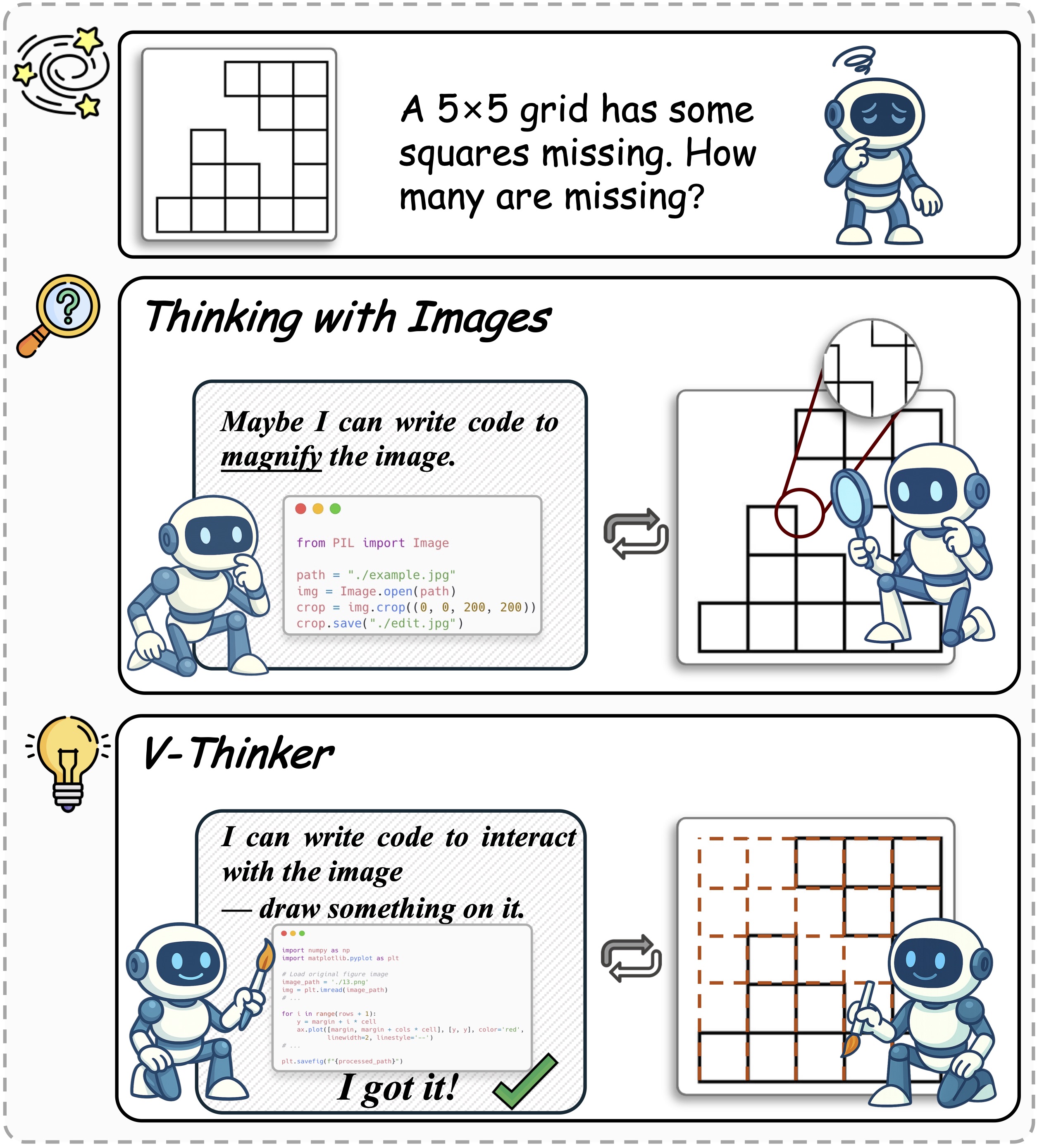}
    }
    \caption{The three paradigms of vision-centric reasoning.}
    \label{fig:intro}
\end{figure}

Humans often simplify complex problem-solving via heuristic multi-modal interactions, especially in vision‑centric reasoning~\citep{tversky2009thinking,Interactive_Visual_Reasoning}. In challenges like geometric proofs, people interactively add auxiliary lines or sketches~\citep{goel1995sketches,tversky2003sketches}, modeling visual relations to solve geometric problems more intuitively. In recent years, visual language models (LMMs) demonstrate exceptional performance in stepwise reasoning tasks, fostering expectations for their development into interactive, image-oriented thinking~\citep{r1onevision,r1v,k1.5,kimivl,seed1.5vl,MM-Eureka,openvlthinker,rl_vllm}. However, despite producing lengthy and coherent chain‑of‑thought (CoT), these models often detach from visual grounding, leading to hallucinations~\citep{Hallu_survey,self-reward-vl,Hallusionbench}. This indicates that current visual reasoning depends more on linguistic priors than on visual perception.

To address these challenges, OpenAI’s o3 model first actively interacts with images during reasoning via visual tools (e.g., cropping, rotation), shifting the paradigm from vision‑assisted reasoning to vision‑centric thinking~\citep{o3,su2025thinking}. Building on this milestone, recent efforts attempt to reproduce o3‑like interactive thinking through end‑to‑end visual reinforcement learning~\citep{deepeye,simpleo3}. 
Moreover, Thyme~\citep{thyme} autonomously generates executable code to render diverse visual operations on images throughout reasoning. However, the available visual actions of them are still limited and heavily depend on precise spatial localization. To further expand the visual tool space and broaden interactive thinking patterns, foundational works such as DeepSketcher~\citep{GeoSketch,zhang2025deepsketcher} enable LMMs to autonomously add auxiliary lines within images, explicitly modeling logical relations between visual elements. While insightful for vision‑centric reasoning, their tool designs are often tightly coupled to specific task types. Moreover, a heavy reliance on "Image2Code" pipelines for image editing struggles to accurately depict spatial relationships between visual elements and may introduce extra noise~\citep{visual_noise,image_noise}. This raises a fundamental research question: \textit{\textbf{How can we effectively integrate image interaction into the visual chain‑of‑thought process?}}

\begin{figure*}[h!]
    \centering
    \resizebox{\textwidth}{!}{
    \includegraphics{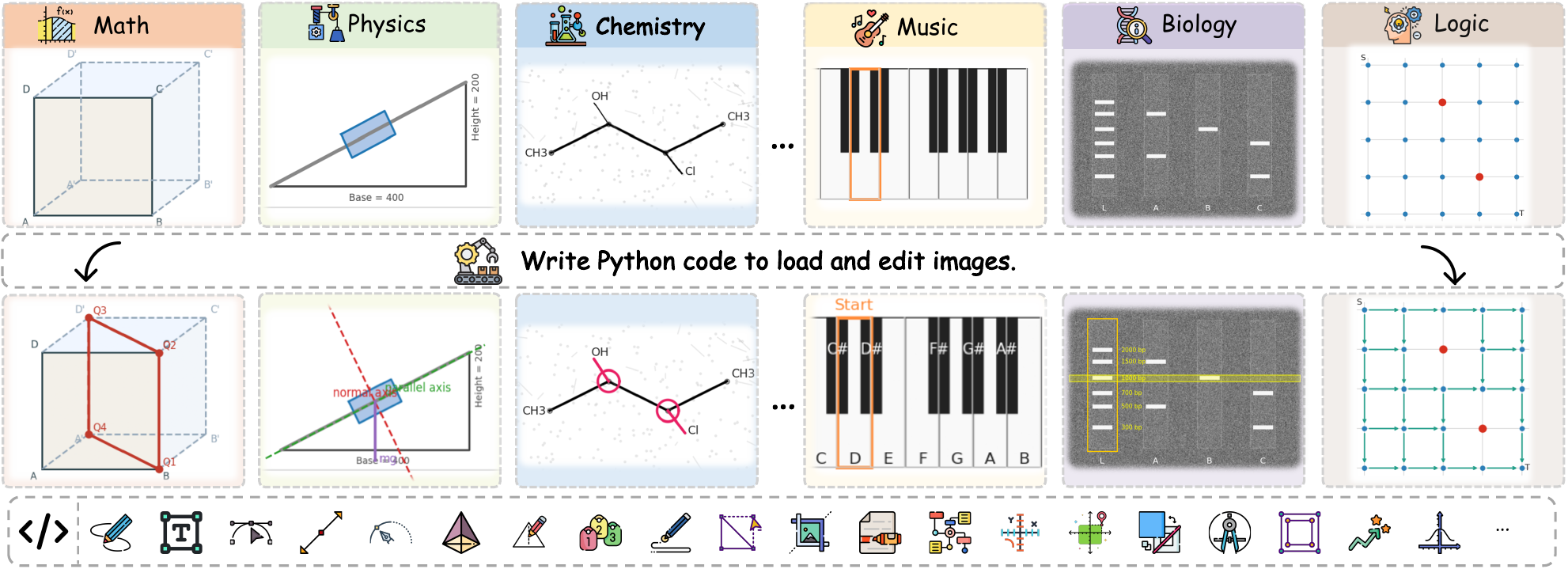}
    }
    \caption{Representative examples of V-Thinker's knowledge-driven synthesis spanning diverse reasoning domains.}
    \label{fig:fig2}
\end{figure*}

In this paper, we introduce \textbf{V-Thinker}, a general‑purpose multimodal reasoning assistant that fosters interactive vision‑centric thinking via end‑to‑end reinforcement training. As shown in Figure~\ref{fig:intro}, V-Thinker aims to simplify complex problems via autonomous interaction with images, thereby advancing the next generation of vision-centric reasoning paradigms. Specifically, V‑Thinker introduces a comprehensive vision-centric post‑training paradigm, comprising two key components: the \textit{"data evolution flywheel"} and a \textit{"progressive interactive training curriculum"} spanning perception to reasoning.

\paragraph{Data Evolution Flywheel.} An ideal vision-centric interactive reasoning model should generalize across real-world diverse tasks. To this end, V‑Thinker targets three fundamental dimensions of interactive reasoning data synthesis: \textbf{\textit{(1) Diversity.}} We directly generate QA pairs using knowledge concepts and visual tool systems, shifting from data expansion to genuine data creation. Then, the flywheel iteratively enlarges both the concept and visual tool sets using newly synthesized data, thereby sustaining a continuous stream of diverse datasets. \textbf{\textit{(2) Quality.}} To ensure strict quality control, we implement a coordinated calibration mechanism where a data checker rigorously screens textual, visual, and image-action dimensions while a repairer calibrates annotations across modalities, ensuring high fidelity. \textbf{\textit{(3) Difficulty.}} Building on the preceding stages, we further introduce a progressive expansion stage that enriches the difficulty ladder by incorporating parallel and sequential extension strategies. 

Through a three-stage data synthesis flywheel, we obtain a high-quality visual interactive reasoning dataset, namely \textit{\textbf{V-Interaction-400K}}.

\paragraph{Visual Progressive Training Curriculum.} Following the data evolution flywheel, we design a two-stage curriculum that progressively builds robust perception and then aligns it with interactive reasoning capabilities: \textbf{\textit{(1) Perception Alignment.}} We model the visual space through three key dimensions and synthesize perception-specific datasets to enhance the LMM's localization capabilities, named \textit{\textbf{V-Perception-40K}}. Furthermore, the LMM is fine-tuned on this dataset to align the localization and referencing of visual anchors. \textbf{\textit{(2) Interactive Reasoning Alignment.}} Built upon the perceptual foundation, we implement cold-start supervised fine-tuning followed by RL within a sandboxed executor environment, enabling the LMM to autonomously generate executable code that interacts with visual elements and maintains coherent reasoning chains grounded in visual evidence.

To thoroughly evaluate LMMs' visual-interactive reasoning capabilities, we introduce \textbf{VTBench}, a benchmark targeting tasks that inherently demand visual interaction. Each instance is verified by domain experts and sourced from diverse public datasets. In summary, the key contributions are as follows:

\begin{itemize}[leftmargin=1em]
\item We formalize \textbf{Interactive Thinking with Images} and introduce \textbf{V‑Thinker}, an end‑to‑end multimodal reasoner that bridges visual grounding and interactive thinking through code‑driven visual tools.


\item We propose a \textbf{Data Evolution Flywheel} that automatically synthesizes, evolves, and verifies interactive reasoning datasets across three dimensions: \textit{Diversity, Quality, and Difficulty}, and further release a large-scale, high-quality visual interaction dataset, \textbf{V-Interaction-400K}, which can also be extended to image-to-code and other vision-language tasks.

\item We introduce a \textbf{Visual Progressive Training Curriculum} that first aligns perception via point-level supervision using a high-quality dataset \textbf{V-Perception-40K}, and then aligns interactive reasoning through a two-stage curriculum training framework.

\item We introduce \textbf{VTBench}, an expert-reviewed benchmark featuring standardized protocols. Extensive experiments show that V-Thinker consistently outperforms mainstream LMM-based baselines on both interactive and general reasoning scenarios, providing valuable insights for advancing image-interactive reasoning.
\end{itemize}

\section{Related Work}
\label{relatedwork}

\paragraph{Multimodal Reasoning.} Recent advances in large language models (LLMs) and multimodal large language models (MLLMs) have significantly enhanced their reasoning capabilities across diverse tasks~\citep{liu2024improved,guo2025seed1,bai2025qwen2,wang2025internvl3,hong2025glm,mcts,team2025kwai,coreteam2025mimovltechnicalreport,an2025llava,voracle,wu2024deepseek,ocrcritic,Qiao_2024_CVPR,autoif,dong2025aepo,dong2025tool,dong2025arpo,dong2024toward,dpa,song2024cs}. Recent efforts such as MathVista~\citep{lu2023mathvista}, MathVision~\citep{wang2024measuring}, MathVerse~\citep{zhang2024mathverse}, We-Math~\citep{qiao2024we}, Dynamath~\citep{zou2024dynamath}, LogicVista~\citep{xiao2024logicvista}, and VisuLogic~\citep{xu2025visulogic} have introduced comprehensive benchmarks that systematically evaluate model performance across mathematical, logical, and visual reasoning scenarios.
Methodologically, prior studies have improved visual–textual alignment~\citep{shi2024math,zhang2024mavis,wang2025mathcoder}, incorporated step-wise reasoning~\citep{zhuang2025math,luo2025ursa}, and explored RL–based optimization to strengthen multimodal reasoning~\citep{huang2025vision,zhang2025r1,meng2025mm,chen2025sft,liu2025noisyrollout,ai2025m2,wan2025srpo,zheng2025deepeyes,yang2025wethink,qiao2025we}. In particular, RL-based frameworks such as MM-Eureka~\citep{MM-Eureka} and Vision-R1~\citep{huang2025vision} have introduced reinforcement learning into visual reasoning, revealing new possibilities for enhancing model reasoning depth. These works have laid a foundation for the continued advancement of visual reasoning.

\paragraph{Thinking with Images.}

Interactive visual reasoning is a long-term research vision. Early explorations, such as LLaVA-Plus~\citep{liu2024llava} and Visual Sketchpad~\citep{hu2024visual}, pioneered this direction by enabling models to conduct visual operations during reasoning. These studies laid an important foundation for integrating interactive perception into multimodal reasoning systems. With the growing success of reinforcement learning (RL) in visual reasoning, OpenAI's o3~\citep{o3} introduced the concept of thinking with images. Subsequent works~\citep{li2025mathopeval,wei2025geoint,wu2025vtool,zhou2025perception,zhang2025deepsketcher,deng2025toolscope,zhou2025reinforced,yuan2025video,li2025look,zhao2025pyvision,shi2025mathcanvas} have further developed this paradigm, provides a fresh perspective for the field of agentic RL. In particular, DeepEyes~\citep{deepeye} and Thyme~\citep{zhang2025thyme} employ RL-based training to guide models in performing executable visual tools (e.g., cropping) as part of their reasoning chain. Meanwhile, DeepSketcher~\citep{zhang2025deepsketcher} explores implicit visual reasoning, where models leverage abstract visual cues instead of explicit pixel-level manipulation. Building on this progress, our work advances interactive thinking with images by enabling the model to autonomously generate, execute, and iteratively refine visual code during reasoning. This opens a new perspective on integrating visual interaction into reasoning, paving the way toward more intuitive and human-like multimodal cognition.

\section{Preliminary}

\subsection{Problem Formulation}
\label{problem formu}
\textbf{Interactive Thinking with Images.} We envision an ideal vision-centric reasoning paradigm where reasoning unfolds through progressive interaction with the image—by perceiving, modifying, and reflecting on its visual state. 
Building upon this idea, \textbf{V-Thinker} treats reasoning as a \emph{code-driven visual interaction process}. 
At each reasoning step, the model generates a textual thought $r_t$ and, when necessary, a code segment $c_t$ that operates on the current image $I_t$. 
The environment $\mathcal{E}$ executes $c_t$, resulting in an updated image $I_{t+1}$. 
Formally,
\begin{equation*}
\resizebox{\linewidth}{!}{$
\mathcal{F}: (Q, I_0) \mapsto (R, A), \quad 
R = \{(r_t, c_t, I_{t+1})\}_{t=1}^{T}, \quad 
I_{t+1} = \mathcal{E}(I_t, c_t)
$}
\end{equation*}
where $Q$ is the task query, $R$ the reasoning trajectory, and $A$ the final answer. 
Through this interactive process, the model reasons by generating code to modify the image and leveraging the resulting visual feedback to guide subsequent reasoning.

\subsection{Rethinking on Data Synthesis Paradigm}
\label{rethinking}

Traditional vision-centric reasoning datasets are built upon manually defined tasks, where models act as solvers generating chain-of-thought reasoning and answers. This data synthesis paradigm limits diversity and scalability, while interactive reasoning requiring precise spatial and logical alignment remains challenging.

With the growth of modern model generation capabilities, we re-examine this paradigm from a new perspective:
\begin{quote}
\noindent \emph{In data synthesis, can models evolve from solving problems to creating problems?}
\end{quote}

To end this, we explore this paradigm from two directions. 

\textbf{(1) Role Transformation: from \textit{"solver"} to \textit{"creator"}}  Traditional data synthesis methods employ models as solvers to distill their reasoning trajectories, based on the core assumption that models cannot directly generate high-quality multimodal QA pairs, particularly those involving complex geometric shapes and spatial relationships. However, this approach constrains the diversity and solvability of synthesized samples. 

\begin{figure}[t!]
    \centering
    \resizebox{0.48\textwidth}{!}{
    \includegraphics{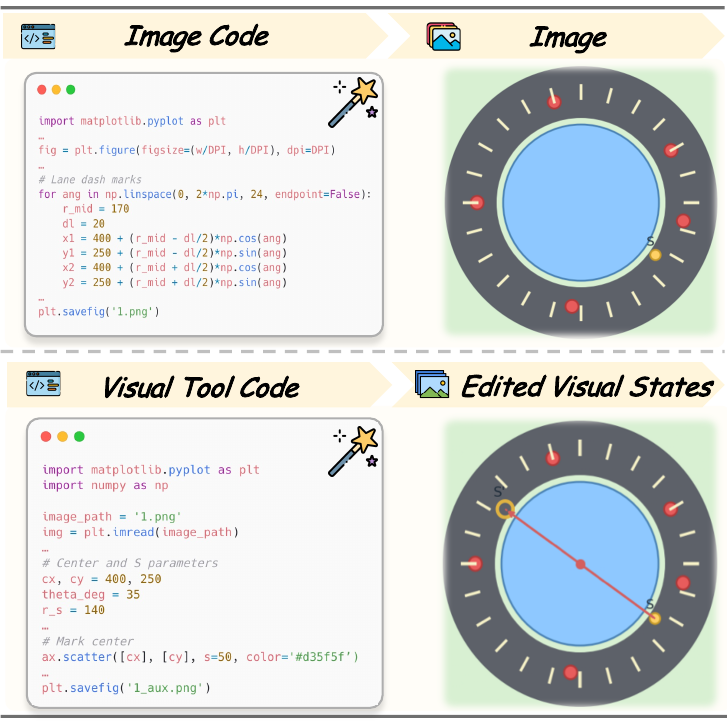}
    }
    \caption{The rendering process from code to image.}

    \label{fig:preprocess}
\end{figure}

In this work, we revisit the role of vision-language models in data synthesis, revealing \textbf{that existing strong LMMs are capable of reaching the creator level in data synthesis.} As shown in Figure~\ref{fig:preprocess}, GPT-5 can directly generate Python code to render high-quality original images along with corresponding auxiliary line diagrams and reasoning trajectories.

Figure~\ref{fig:fig2} shows representative rendered image comparisons. The synthesis quality far exceeds our initial expectations. The exceptional coding capabilities enable fine-grained, outstanding image editing, such as highlighting chemical elements and annotating vertical guide lines, directly creating image annotations. Therefore, we shift from the distillation paradigm, enabling innovative generation of complex problems, diagrams, and reasoning trajectories rather than passively responding to predefined tasks.

\textbf{(2) Knowledge-Driven Representation.} Knowledge concepts serve as anchors for human knowledge, where an ideal comprehensive knowledge system can generate reasoning data covering diverse scenarios. We therefore move beyond using seed samples as references and instead employ structured knowledge systems, where \textbf{knowledge concepts serve as condensed representations of reasoning semantics, capturing diverse real-world scenarios.} Knowledge specifies \textit{what} to reason about, while models determine \textit{how} to unfold reasoning through interaction. As shown in Figure~\ref{fig:fig2}, by providing only knowledge concepts and their descriptions, knowledge-driven synthesis produces problems with precise spatial alignment, coherent reasoning, and diverse visual structures, broadening the construction and evolution of reasoning data.

\begin{figure*}[ht!]
    \centering
    \resizebox{\textwidth}{!}{
    \includegraphics{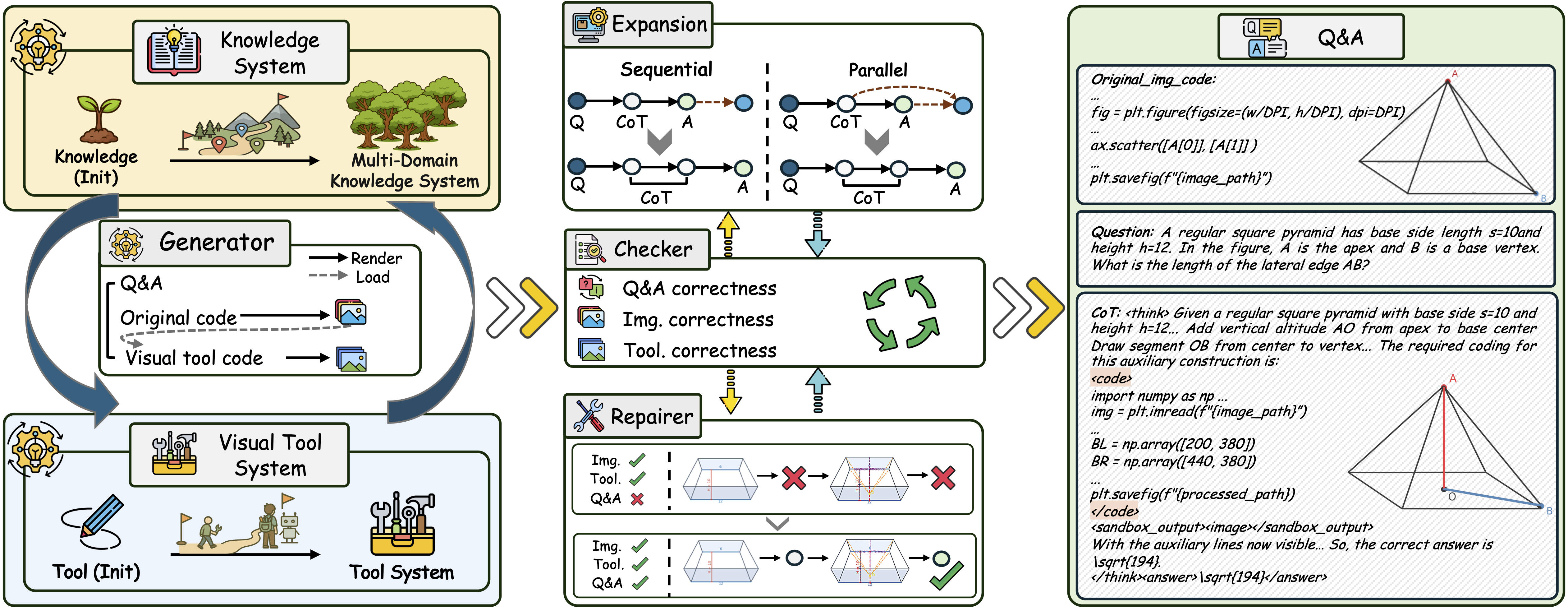}
    }
    \caption{The Data Evolution Flywheel framework: \textbf{Left:} knowledge-driven evolution mechanism. \textbf{Middle:} coordinated calibration and progressive expansion stages. \textbf{Right:} representative synthetic QA instances generated through the flywheel.}

    \label{fig:data}
\end{figure*}

\section{Methodology}
\label{Methodology}

\paragraph{Overview.} This section introduces V-Thinker, a general-purpose multimodal reasoning assistant that enables interactive vision-centric reasoning through end-to-end reinforcement learning. As shown in Figure~\ref{fig:data}, V-Thinker comprises two core components:

\begin{enumerate}[leftmargin=1em]
\item \textbf{Data Evolution Flywheel (§\ref{sec:evolution flyheel}):} We automatically synthesize, evolve, and verify interactive reasoning datasets across three dimensions: Diversity, Quality, and Difficulty.
\item \textbf{Visual Progressive Training Curriculum (§\ref{sec:visual_training}):} We introduce a two-stage training framework to achieve progressive alignment from perception understanding to vision-centric interactive reasoning patterns.
\end{enumerate}
Below, we delve into the specifics of our approach.

\subsection{Data Evolution Flywheel}
\label{sec:evolution flyheel}

Building on the knowledge-driven paradigm outlined in §~\ref{rethinking}, we design the \textbf{Data Evolution Flywheel}, an automated, scalable, and verifiable framework for synthesizing interactive reasoning data. The framework is divided into three processes (as shown in Figure~\ref{fig:data}):

\begin{itemize}[leftmargin=1em, topsep=2pt, itemsep=2pt]
\item \textbf{Knowledge-driven Evolution} (§\ref{sec:generator}) generates reasoning data through the co-evolution of knowledge and tool sets, iteratively expanding both to yield diverse problems and reasoning trajectories.

\item \textbf{Coordinated Calibration} (§\ref{sec:checker}) verifies the correctness of generated questions, rendered images, and edited visual states, ensuring consistency across text, code execution, and visual outcomes.

\item \textbf{Progressive Expansion} (§\ref{sec:explorer}) enhances reasoning difficulty via iterative step extension and compositional knowledge integration, gradually constructing more complex and challenging reasoning chains.
\end{itemize}

\subsubsection{Diversity: Knowledge-driven Evolution}
\label{sec:generator}

Prior data synthesis methods rely on curated seed images, inherently constraining diversity. We propose that \textit{\textbf{foundational knowledge concepts and visual tool system serve as the most granular anchors for synthesizing interactive data, enabling orthogonal and diverse generation.}} Our pipeline avoids distillation and instead instructs a strong LMM to create datasets from scratch.

Following this guideline, given an initial knowledge system $\mathcal{K}_0$ derived from \textit{We-Math 2.0}~\citep{qiao2025we} and a curated task-centric tool set $\mathcal{T}_0$, knowledge and tools jointly drive iterative data synthesis through a co-evolutionary loop (Algorithm~\ref{alg:constructive}).

At each iteration $n$, sampled combinations $\mathsf{Combos}(\mathcal{K})$ from the current knowledge system are provided to the strong generator $\mathcal{G}$ (e.g. GPT5) to construct reasoning data:
\begin{equation}
(\mathcal{D}_K, \hat{\mathcal{T}}) = \mathcal{G}(\mathsf{Combos}(\mathcal{K})).
\end{equation}

Each generated instance contains a question $Q$, the original code $c_0$ that renders the problem-specific image $I_0$, the corresponding tool predictions $\hat{\mathcal{T}}$, and a reasoning trajectory $R = \{(r_t, c_t, I_{t+1})\}_{t=1}^{T_R}$, where each executable code $c_t$ renders the visual component $I_{t+1}$. Figure~\ref{fig:preprocess} illustrates the image rendering process from code to image.

In parallel, tool-driven generation applies the same procedure on $\mathsf{Combos}(\mathcal{T})$, producing data $\mathcal{D}_T$ and predicted knowledge $\hat{\mathcal{K}}$:
\begin{equation}
(\mathcal{D}_T, \hat{\mathcal{K}}) = \mathcal{G}(\mathsf{Combos}(\mathcal{T})).
\end{equation}

Together, these complementary processes constitute a co-evolutionary loop in which $\mathcal{K}$ and $\mathcal{T}$ continuously generate new counterparts. The newly predicted elements are then integrated via expansion functions $\Phi_K$ and $\Phi_T$:
\begin{equation}
\mathcal{K} \leftarrow \mathcal{K} \cup \Phi_K(\hat{\mathcal{K}}, \mathcal{K}), \qquad 
\mathcal{T} \leftarrow \mathcal{T} \cup \Phi_T(\hat{\mathcal{T}}, \mathcal{T}),
\end{equation}

where $\Phi_K$ and $\Phi_T$ filter, merge, and normalize novel elements during incorporation via BGE-based hierarchical clustering. The co-evolution mechanism among synthetic data, knowledge concepts, and visual tool systems is illustrated in the Figure~\ref{fig:data} (left).

Through repeated execution over $N$ rounds, the system gradually enriches both $\mathcal{K}$ and $\mathcal{T}$, ultimately yielding an initial dataset $\mathcal{D}_{\text{init}}$ that serves as the foundation for subsequent calibration and expansion stages.

\begin{algorithm}[t]
\small
\caption{Constructive Evolution for Dataset Synthesis}
\label{alg:constructive}
\begin{algorithmic}[1]
\Require Initial knowledge set $\mathcal{K}_0$, tool set $\mathcal{T}_0$, generator $\mathcal{G}$, expansion rules $\Phi_K, \Phi_T$, iterations $N$
\Ensure Evolved knowledge $\mathcal{K}^\ast$, tools $\mathcal{T}^\ast$, and dataset $\mathcal{D}_{\text{init}}$

\State $\mathcal{K} = \mathcal{K}_0$, $\mathcal{T} = \mathcal{T}_0$, $\mathcal{D}_{\text{init}} = \emptyset$\Comment{Initialize from scratch}
\For{$n = 1, 2, \dots, N$}
    \State $(\mathcal{D}_K, \hat{\mathcal{T}}) \gets \mathcal{G}(\mathsf{Combos}(\mathcal{K}))$ \Comment{Generate QA and predicted tools}
    \State $(\mathcal{D}_T, \hat{\mathcal{K}}) \gets \mathcal{G}(\mathsf{Combos}(\mathcal{T}))$ \Comment{Generate QA and predicted knowledge concepts}
    \State $\mathcal{D}_{\text{init}} \gets \mathcal{D}_{\text{init}} \cup \mathcal{D}_K \cup \mathcal{D}_T$ \Comment{Accumulate new samples}
    \State $\Delta\mathcal{K} \gets \Phi_K(\hat{\mathcal{K}}, \mathcal{K})$ \Comment{Knowledge expansion}
    \State $\Delta\mathcal{T} \gets \Phi_T(\hat{\mathcal{T}}, \mathcal{T})$ \Comment{Tool expansion}
    \State $\mathcal{K} \gets \mathcal{K} \cup \Delta\mathcal{K},\quad \mathcal{T} \gets \mathcal{T} \cup \Delta\mathcal{T}$
\EndFor
\State \textbf{Output:} $\mathcal{K}^\ast = \mathcal{K},\ \mathcal{T}^\ast = \mathcal{T},\ \mathcal{D}_{\text{init}}$
\end{algorithmic}
\end{algorithm}

\subsubsection{Quality: Coordinated Calibration}
\label{sec:checker}

After obtaining the synthesized diverse dataset $\mathcal{D}_{\text{init}}$, strict quality control is essential. Therefore, we introduce a regulative calibration stage to ensure multi-level consistency across generated samples. This stage consists of two modules.

\textbf{Checker.} As shown in Figure~\ref{fig:data} (Mid), each instance is examined by a data checking module $\mathcal{V}$ that verifies (1) answer correctness, (2) validity of the rendered original image, and (3) coherence of intermediate visual states produced during reasoning. Only samples satisfying all three criteria are retained as valid candidates.

\textbf{Repairer.} For cases where the textual answer is incorrect but the rendered image is valid and intermediate visual states remain coherent, we follow the principle that \textit{\textbf{a reasoning chain is fundamentally guided by its question}}, reconstructing the question from original and edited visual states to realign textual and visual reasoning. The reconstructed instances are re-evaluated by $\mathcal{V}$, and the loop repeats until inconsistencies are resolved.

Through the iterative calibration of data checking and repair, we refine $\mathcal{D}_{\text{init}}$ into a coherent and verified dataset $\mathcal{D}_{\text{verified}}$, which serves as the foundation for subsequent progressive expansion.

\subsubsection{Difficulty: Progressive Expansion}
\label{sec:explorer}

To establish difficulty-stratified data partitions and deepen reasoning chain complexity, our intuition is to extend the CoT's context length to achieve difficulty escalation, thereby inversely mapping more challenging QA pairs. This naturally motivates us to introduce two complementary strategies to progressively expand the difficulty: \textbf{parallel} and \textbf{sequential} extensions, constructing more challenging QA pairs:

\textbf{(1) Parallel Extension.} New auxiliary constructions are introduced independently of the existing ones, providing additional key observations that complement the original reasoning to reach the final answer.

\textbf{(2) Sequential Extension.} New auxiliary constructions are closely linked to the existing reasoning, requiring prior results or original geometric entities to define subsequent operations. For example, if the original auxiliary line is $DM$, a new line perpendicular to $DM$ can be introduced to support further deductions.

\begin{figure}[t!]
    \centering
    \resizebox{0.47\textwidth}{!}{
    \includegraphics{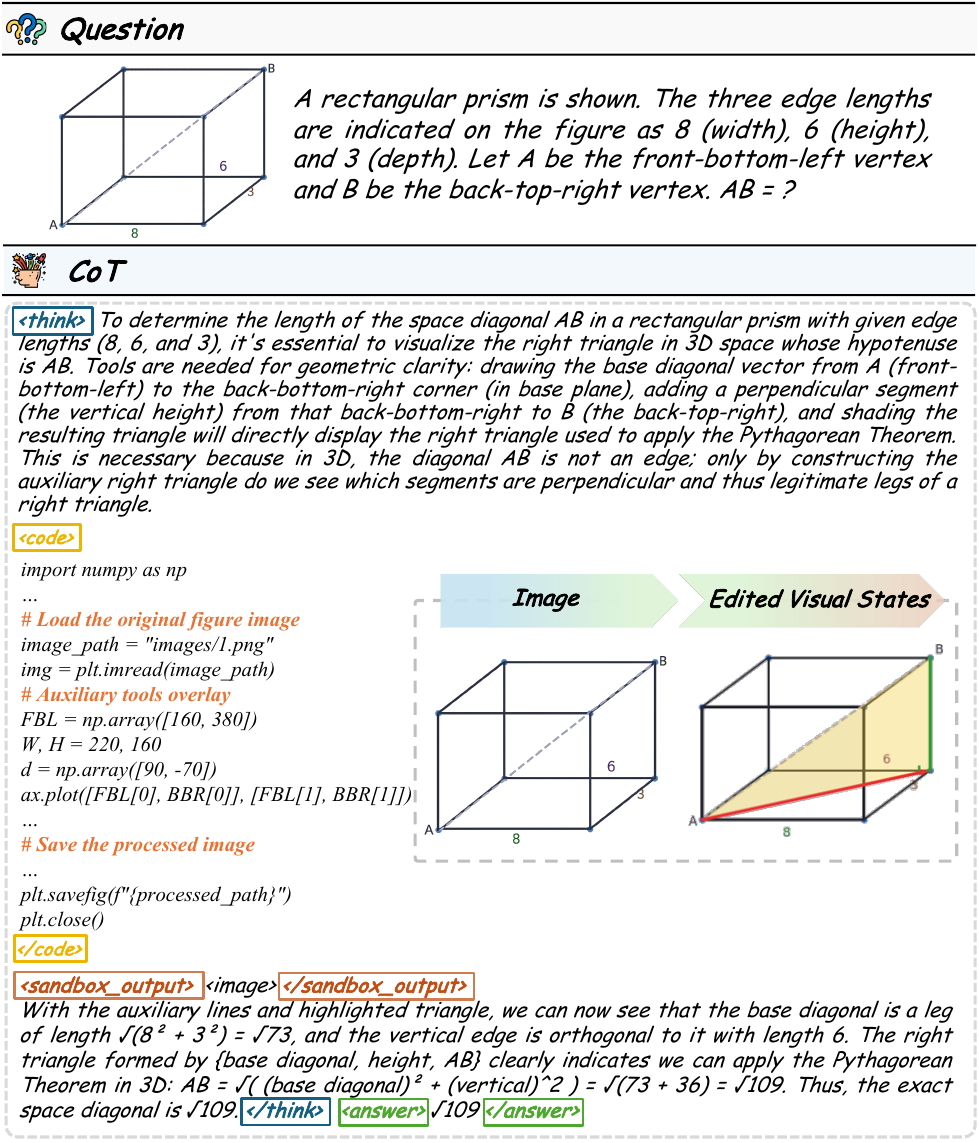}
    }
    \caption{A representative sample from the synthesized dataset V-Interaction-400K ($\mathcal{D}$).}

    \label{fig:dataset_pre}
\end{figure}

Under these two strategies, a subset of verified data $\mathcal{D}_{\text{verified}}$ is sampled and provided to the expansion model, which generates extended reasoning code segments and corresponding visual states. The generated extensions are re-validated by the verification module $\mathcal{V}$ and iteratively refined until convergence. We limit the maximum extension depth to three steps and merge all results to form the final dataset $\mathcal{D}$. Figure~\ref{fig:dataset_pre} display a representative sample from the synthesized dataset $\mathcal{D}$.

\subsection{Visual Progressive Training Curriculum}
\label{sec:visual_training}

For reliable vision-centric interactive reasoning, accurate perception of visual elements is crucial. While recent multimodal models demonstrate strong reasoning ability, they often struggle with fine‑grained spatial perception, failing to precisely localize points, intersections, and other anchors. 

To address this, we develop a visual progressive training curriculum that aligns perception and interaction, starting with perceptual grounding through point‑level supervision (§~\ref{sec:percep align}) and advancing to interactive reasoning via progressive alignment training (§~\ref{sec:tool align}).

\subsubsection{Perception Alignment}
\label{sec:percep align}

\paragraph{Perception Data Synthesis.} A critical aspect of perception training is synthesizing perception-specific data. Following the paradigm in Section~\ref{rethinking}, we model visual space via three dimensions: \textit{element relations, element count, and knowledge concepts}. These aspects jointly define visual information, with style governed by knowledge and complexity by element count.

\begin{itemize}[leftmargin=1em, topsep=2pt, itemsep=2pt]
\item \textbf{Element Relations ($P_{\mathcal{E}}$):} We model visual elements such as points, lines, angles, and circles, based on foundational geometry principles~\citep{fitzpatrick2008euclid}, and extend this set with textual and symbolic elements for alignment with real-world scenarios. This modeling integrates both element types and their spatial relationships, such as "point on line" and "point outside circle," capturing the core geometric interactions in visual reasoning.

\item \textbf{Element Count ($P_{\mathcal{C}}$):} We sample the number of elements from a normal distribution $\mathcal{N}(\mu=8, \sigma^2=4)$ to control task complexity, where $\mu$ represents the mean and $\sigma^2$ the variance.

\item \textbf{Knowledge Concepts ($P_{\mathcal{K}}$):} We sample from the knowledge system $\mathcal{K}$ to define reasoning objectives for each task.
\end{itemize}

\begin{figure}[t!]
    \centering
    \resizebox{0.47\textwidth}{!}{
    \includegraphics{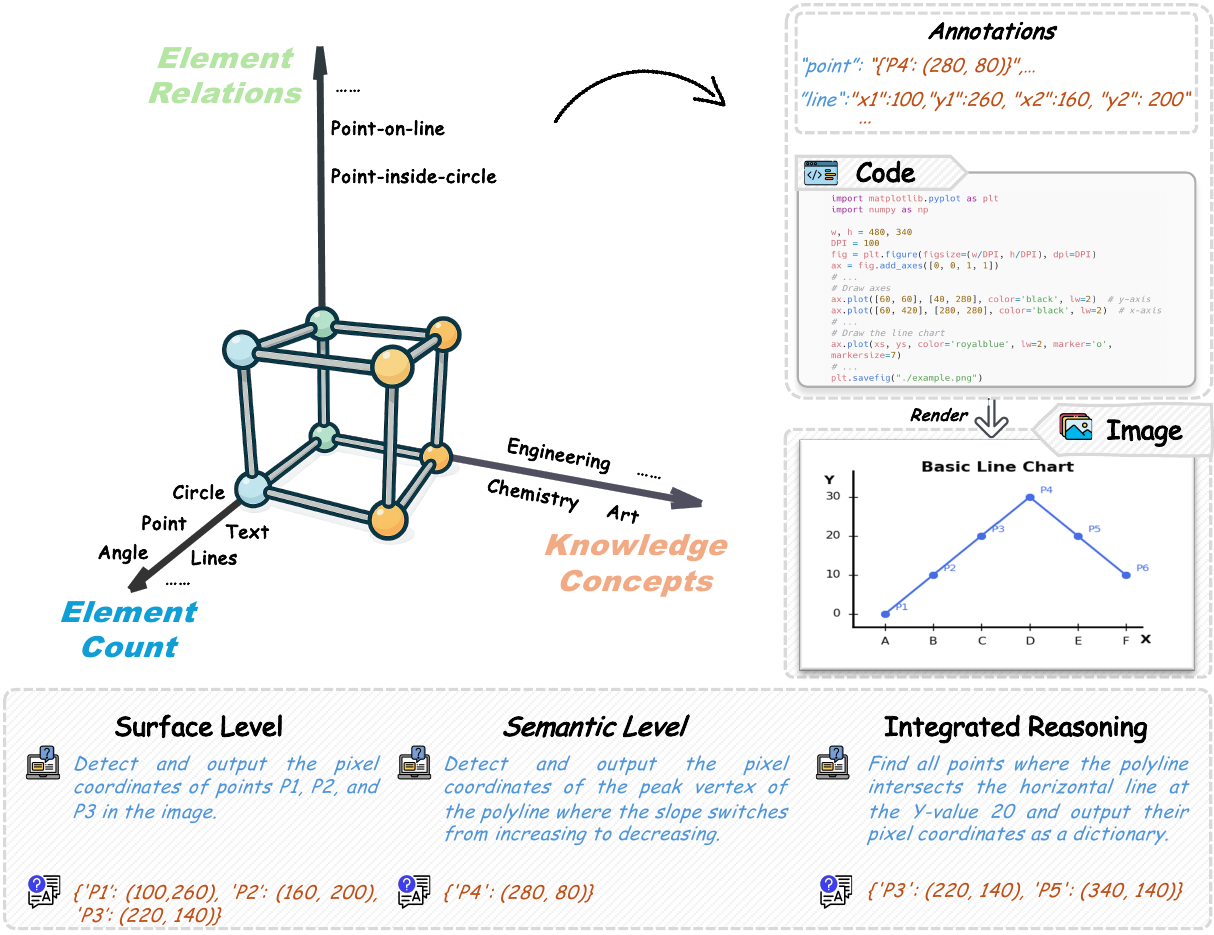}
    }
    \caption{The overview of the perception data synthesis.}

    \label{fig:pecep}
\end{figure}

As shown in Figure~\ref{fig:pecep}, each combination of these dimensions generates corresponding coordinates, which serve as visual tags representing the spatial relationships between elements. Using these coordinates, we then structure the tasks into three levels of complexity: surface-level perception, semantic-level reasoning, and integrated reasoning:

\begin{itemize}[leftmargin=1em, topsep=2pt, itemsep=2pt]
\item \textbf{Surface-level perception:} Basic tasks, such as identifying the coordinates of a specific point, for example, point~$A$.
\item \textbf{Semantic-level reasoning:} Tasks requiring geometric understanding, such as identifying the top-left vertex of a cube.
\item \textbf{Integrated reasoning:} Tasks combining perception and computation, such as finding the center of a cube based on its dimensions.
\end{itemize}

This hierarchical structure generates diverse question-answer pairs, forming the dataset $\mathcal{D}_{\text{perception}}$, which spans all complexity levels and enhances perceptual capabilities.

\paragraph{Perception Training.} The model is trained using supervised fine-tuning, with the objective of minimizing the loss function:
\begin{equation}
\mathcal{L}_{\text{SFT}}(\theta) = \mathbb{E}_{(Q, A) \sim \mathcal{D}_{\text{perception}}} \left[ -\log P_\theta(A \mid Q) \right]
\end{equation}
This process enhances the model's ability to process visual information, particularly for tasks involving point-level localization.

\subsection{Interactive Reasoning Alignment}
\label{sec:tool align}

\paragraph{Data Selection.} For comprehensive training, we select data from the following sources:

\textit{\textbf{SFT Dataset:}} In the Cold-Start Fine-tuning stage, we first construct the $D_{\text{perception}}$ dataset to fine-tune V-Thinker with fine-grained perception capabilities, and then align $D$ to equip it with basic interactive reasoning abilities.

\textit{\textbf{RL Dataset:}} To progressively align V-Thinker for robust interactive reasoning, we consider the following two components of the dataset: \textit{\textbf{(1) Open-Source Samples:}} We use the We-Math 2.0~\citep{qiao2025we}, MMK12~\citep{MM-Eureka} and ThinkLite~\citep{wang2025sota}, which provide a broad range of visual reasoning data, from basic to complex, ensuring comprehensive coverage of visual reasoning tasks. \textit{\textbf{(2) Targeted Sampling from $\mathcal{D}$:}} We select data from $\mathcal{D}$ where the base model gives incorrect answers for the original image but correct answers for the edited version. A total of 3,000 instances are sampled from $\mathcal{D}$, ensuring a mix of simpler and more complex data, aimed at improving the model's reasoning ability with visual states.

\paragraph{Cold-Start Fine-tuning}
\label{sec:coldstart}

To enable the model to generate code and interact with visual elements, we apply standard supervised fine-tuning objective on the dataset $\mathcal{D}$, consisting of input-output pairs $(x_i, y_i)$, where $x_i$ includes both the problem statement and the corresponding visual information, and $y_i$ represents the reasoning chain with Python code and the edited visual elements. The objective is to minimize the loss function:
\begin{equation}
\mathcal{L}_{\text{SFT}}(\theta) = - \mathbb{E}_{(x_i, y_i) \sim \mathcal{D}} \left[ \log P_\theta(y_i \mid x_i) \right]
\end{equation}
This phase initially enables the model to generate code that reads and interacts with images to perform reasoning tasks.

\paragraph{Reinforcement Learning for Interactive Reasoning}
\label{sec:RLRL}

We apply reinforcement learning (RL) to enhance the model's interactive reasoning capabilities. Building on the Thyme~\citep{zhang2025thyme}, we use its sandbox environment to execute the model's generated code. The model decodes reasoning tasks into executable code, which interacts with the visual elements, and the resulting outputs are fed back into the reasoning process. For optimization, we adopt Group Relative Policy Optimization (GRPO)~\citep{shao2024deepseekmath}, which has been shown to be effective for diverse tasks. Given an input question $x$ and a policy model $\pi_\theta$, GRPO enables the reference policy $\pi_{\text{ref}}$ to generate a group of $G$ outputs $\{y_1, y_2, \ldots, y_G\}$ and optimizes the policy by maximizing:
\begin{multline}
\mathcal{L}_{\text{RL}}(\theta)=\mathbb{E}_{x \sim \mathcal{D}}\Bigg[\frac{1}{\sum_{j=1}^{G} T_{j}} \sum_{j=1}^{G} \sum_{t=1}^{T_{j}} \min \Big(\delta_{j,t} \tilde{A}^{(t)}, \\
\operatorname{clip}\left(\delta_{j,t}, 1-\epsilon_{l}, 1+\epsilon_{h}\right) \tilde{A}^{(t)}\Big)\Bigg],
\end{multline}

where $\delta_{j,t} = \frac{\pi_{\theta}(y_{j,t} \mid x, y_{j,<t})}{\pi_{\text{ref}}(y_{j,t} \mid x, y_{j,<t})}$ represents the importance sampling ratio for the $t$-th token of the $j$-th output, and $\tilde{A}^{(t)}$ denotes the advantage at time step $t$.

\paragraph{Reward Design.} 
We follow a reward function based on the Thyme framework~\citep{zhang2025thyme}, consisting of three components: accuracy ($R_{\text{acc}}$), formatting ($R_{\text{format}}$), and tool usage ($R_{\text{tool}}$). 
The total reward is defined as:
\begin{equation}
\small
R(\tau) = R_{\text{acc}}(\tau) + \lambda_1 \cdot R_{\text{format}}(\tau) + \lambda_2 \cdot \mathbb{I}_{R_{\text{acc}}(\tau) > 0} \cdot R_{\text{tool}}(\tau)
\end{equation}
where $\mathbb{I}_{R_{\text{acc}}(\tau) > 0}$ is an indicator function ensuring that the tool usage reward is applied only when the final answer is correct and involves at least one tool. 
We empirically set $\lambda_1 = 0.5$ and $\lambda_2 = 0.3$ to balance the contributions of formatting and tool usage.

\begin{figure*}[ht!]
    \centering
    \resizebox{\textwidth}{!}{
    \includegraphics{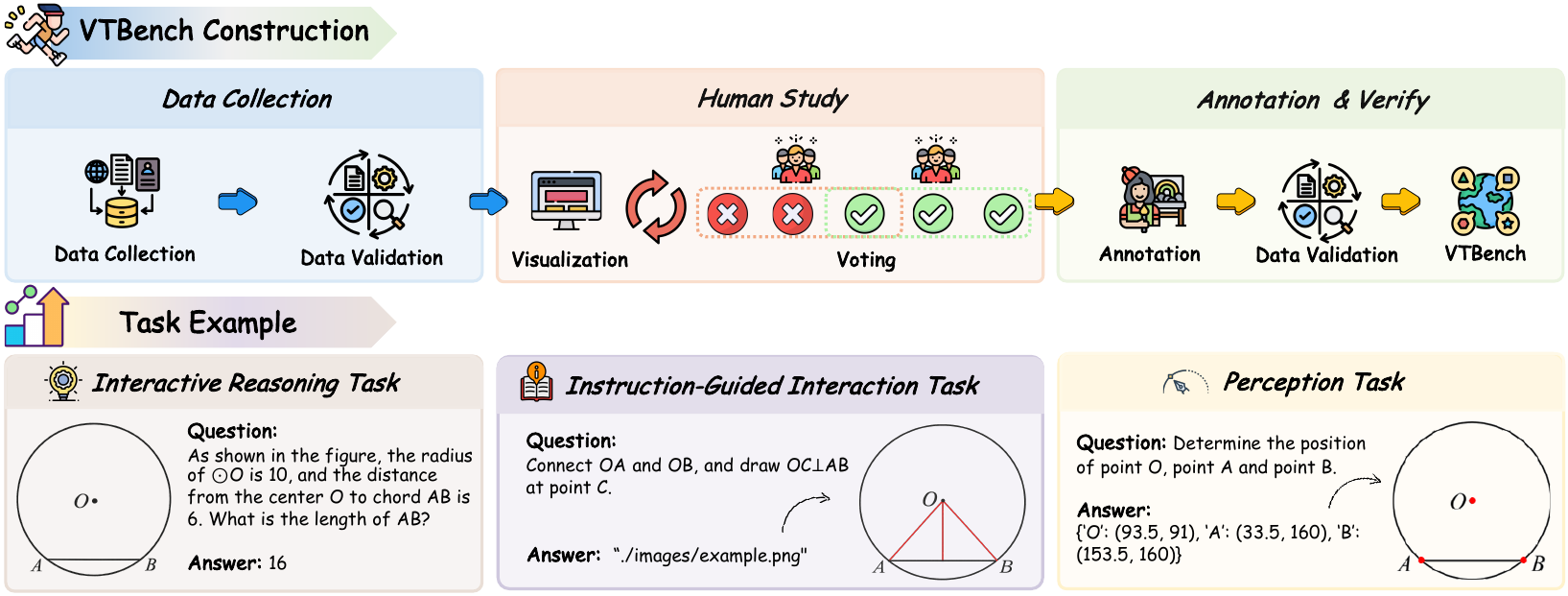}
    }
    \caption{The construction guideline of our VTBench.}

    \label{fig:vtbench}
\end{figure*}

\section{VTBench}
\label{sec:VTBench}

To evaluate the model's vision-centric interactive reasoning capabilities, we introduce VTBench, a benchmark designed for tasks that inherently require interaction with visual elements. The design of VTBench follows the following principles:

\begin{itemize}[leftmargin=1em, topsep=2pt, itemsep=2pt]
\item \textbf{Specific Interactive Tasks:} VTBench prioritizes tasks that inherently require interaction with visual elements, such as drawing auxiliary lines or labeling.

\item \textbf{Expert Evaluation:} Each sample is assessed by a team of five experts, who determine whether visual interaction is necessary for solving the problem. The sample is included if at least three experts agree on its necessity.

\item \textbf{Diversity and Expansion:} To ensure broad coverage, we extend the benchmark by collecting and annotating datasets from multiple sources, including both open-source benchmarks and additional tasks from public platforms, following the same annotation procedure.
\end{itemize}

\subsection{Data Collection and Annotation} As shown in Figure~\ref{fig:vtbench}, VTBench is constructed in two stages:

\begin{itemize}[leftmargin=1em, topsep=2pt, itemsep=2pt]
\item \textbf{Sample Selection:} We collect samples from open-source benchmarks: MathVista~\citep{lu2023mathvista}, MathVision~\citep{wang2024measuring}, MathVerse~\citep{zhang2024mathverse}, Dynamath~\citep{zou2024dynamath}, We-Math~\citep{qiao2024we}, LogicVista~\citep{xiao2024logicvista}, CMM-Math~\citep{liu2024cmm}, CharXiv~\citep{wang2024charxiv} and ZeroBench~\citep{roberts2025zerobench}, as well as additional samples gathered from public platforms. Expert evaluation is then used to determine whether interaction with visual elements is required. Each sample is included if at least three out of five experts agree on its necessity.

\item \textbf{Interaction Annotation:} We design interaction instructions based on the problem-solving chain of thought for each sample. The expert team manually generates interaction graphs in the interactive interface, which simultaneously captures perceptual coordinates. These interaction tags are then transformed into QA pairs with GPT-4.1, followed by expert validation to ensure consistency and accuracy.
\end{itemize}

\subsection{Evaluation Dimensions} VTBench evaluates vision-centric interactive reasoning capabilities across three hierarchical dimensions, modeling the problem-solving process from perception to adaptive interaction during reasoning (in Figure~\ref{fig:vtbench}):
\begin{equation*}
\resizebox{0.48\textwidth}{!}{$\textit{Perception} \rightarrow \textit{Instruction-Guided Interaction} \rightarrow \textit{Interactive Reasoning}$}
\end{equation*}

\begin{itemize}[leftmargin=1em, topsep=2pt, itemsep=2pt]
\item \textbf{Perception:} Tasks that assess the model's ability to recognize and interpret visual elements, such as identifying coordinates.

\item \textbf{Instruction-Guided Interaction:} Tasks where the model receives explicit instructions (e.g., drawing lines or labeling) and must interact with the visual elements to fulfill these instructions.

\item \textbf{Interactive Reasoning:} Tasks that require the model to solve reasoning tasks involving visual interaction, such as drawing auxiliary lines or modifying diagrams.
\end{itemize}

\begin{table*}[t]
\centering
\setlength\tabcolsep{5.2pt}
\renewcommand{\arraystretch}{1.0}
\begin{tabular}{
 p{3.5cm}
c c c c 
c c c c
}
\toprule
\multirow{2}{*}{\textbf{Method}}
& \multicolumn{4}{c}{\textbf{VTBench}}
& \multicolumn{4}{c}{\textbf{General Reasoning}} \\
\cmidrule(lr){2-5} \cmidrule(lr){6-9}
& \textit{Perception}
& \makecell{\textit{Instruct.}\\\textit{Interaction}}
& \makecell{\textit{Interactive}\\\textit{Reasoning}}
& \textit{Avg.}
& \makecell{\textit{MathVision}\\\textit{Acc.}}
& \makecell{\textit{We-Math}\\\textit{Acc.}}
& \makecell{\textit{VisuLogic}\\\textit{Acc.}}
 & \textit{Avg.} \\
\midrule

\rowcolor[RGB]{245,245,250}
GPT-4o & 12.6 & 26.0 & 36.4 & 25.0 & 43.8 & 68.8 & 26.3 & 46.3 \\

InternVL3-78B & 13.8 & 19.0 & 43.4 & 25.4 & 43.1 & 64.2 & 27.7 & 45.0 \\

\rowcolor[RGB]{245,245,250}
InternVL3-8B & 10.4 & 6.8 & 33.8 & 17.0 & 29.3 & 58.8 & 24.9 & 37.7 \\

LLaVA-OV-1.5-8B & 12.2 & 12.2 & 30.2 & 18.2 & 25.6 & 56.7 & 23.7 & 35.3 \\

\rowcolor[RGB]{245,245,250}
InternVL3-2B & 3.0 & 3.4 & 22.0 & 9.5 & 23.3 & 41.7 & 24.3 & 29.8 \\

Qwen2.5-VL-7B & 12.6 & 8.8 & 31.8 & 17.7 & 23.0 & 61.7 & 26.0 & 36.9 \\

\hdashline

\rowcolor{gray!8}
\textbf{V-Thinker-7B}
& \textbf{18.6}
& \textbf{31.6}
& \textbf{40.4}
& \textbf{30.2}
& \textbf{29.3}
& \textbf{62.8}
& \textbf{26.6}
& \textbf{39.6} \\

\rowcolor{gray!8}
$\Delta$ (\textit{vs} Qwen2.5-VL-7B)
& \textbf{\textit{\textcolor{red!50!black}{+6.0}}}
& \textbf{\textit{\textcolor{red!50!black}{+22.8}}}
& \textbf{\textit{\textcolor{red!50!black}{+8.6}}}
& \textbf{\textit{\textcolor{red!50!black}{+12.5}}}
& \textbf{\textit{\textcolor{red!50!black}{+6.3}}}
& \textbf{\textit{\textcolor{red!50!black}{+1.1}}}
& \textbf{\textit{\textcolor{red!50!black}{+0.6}}}
& \textbf{\textit{\textcolor{red!50!black}{+2.7}}} \\

\bottomrule
\end{tabular}
\caption{Overall performance on VTBench (left) and general reasoning (right).
(\textit{Instruct.~Interaction} denotes \textit{Instruction-Guided Interaction}.)}
\label{tab:general_reasoning}
\end{table*}

\subsection{Data Statistics and Evaluation Metrics}

VTBench comprises 1,500 question-answer pairs across three task types, with 500 samples per task. It incorporates 9 open-source benchmarks across four domains (\textit{Logical Reasoning, Geometry, Algebra, Statistics}), categorized into three key evaluation metrics:

\begin{itemize}[leftmargin=1em, topsep=2pt, itemsep=2pt]
\item \textbf{For Perception Task:} Considering the inconsistency in coordinate systems across different models, the model is instructed to generate Python code that draws the point at the perceived location. The generated image, compared with the annotated image, is judged by LMMs.

\item \textbf{For Instruction-Guided Interaction Task:} For tasks that require explicit instructions (e.g., drawing lines or labeling regions), the model is instructed to generate Python code to perform the required visual interaction. The result, compared with the annotated image, is judged by LMMs.

\item \textbf{For Interactive Reasoning Task:} For reasoning tasks, the model generates answers, which are then evaluated by Large Language Models (LLMs) based on correctness.
\end{itemize}


\begin{figure*}[ht]
    \centering
    \resizebox{0.9\textwidth}{!}{
    \includegraphics{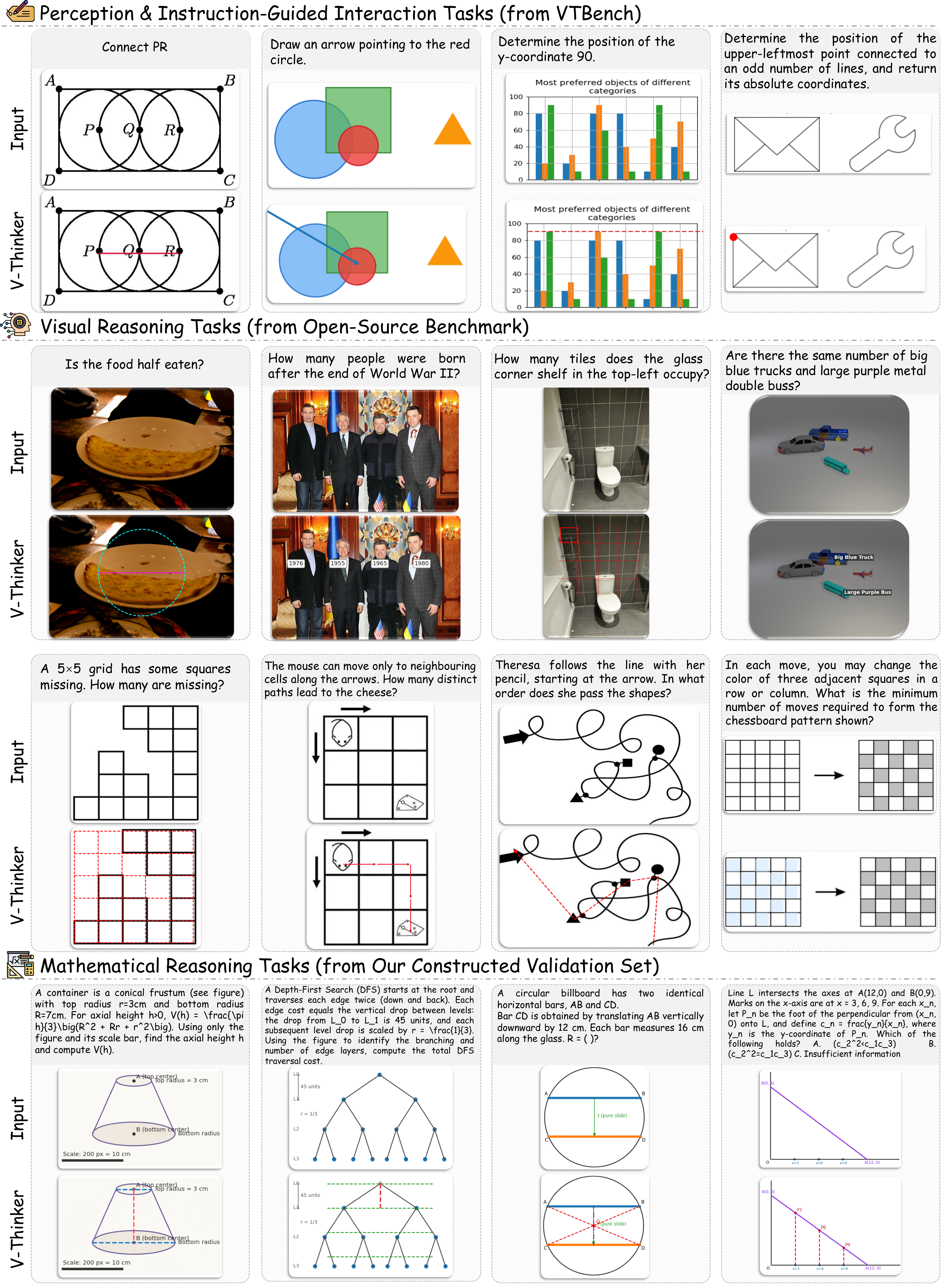}
    }
    \caption{Qualitative analysis of V-Thinker-7B on vision-centric interactive reasoning tasks.}

    \label{fig:caseexp}
\end{figure*}

\begin{figure*}[ht!]
    \centering
    \resizebox{\textwidth}{!}{
    \includegraphics{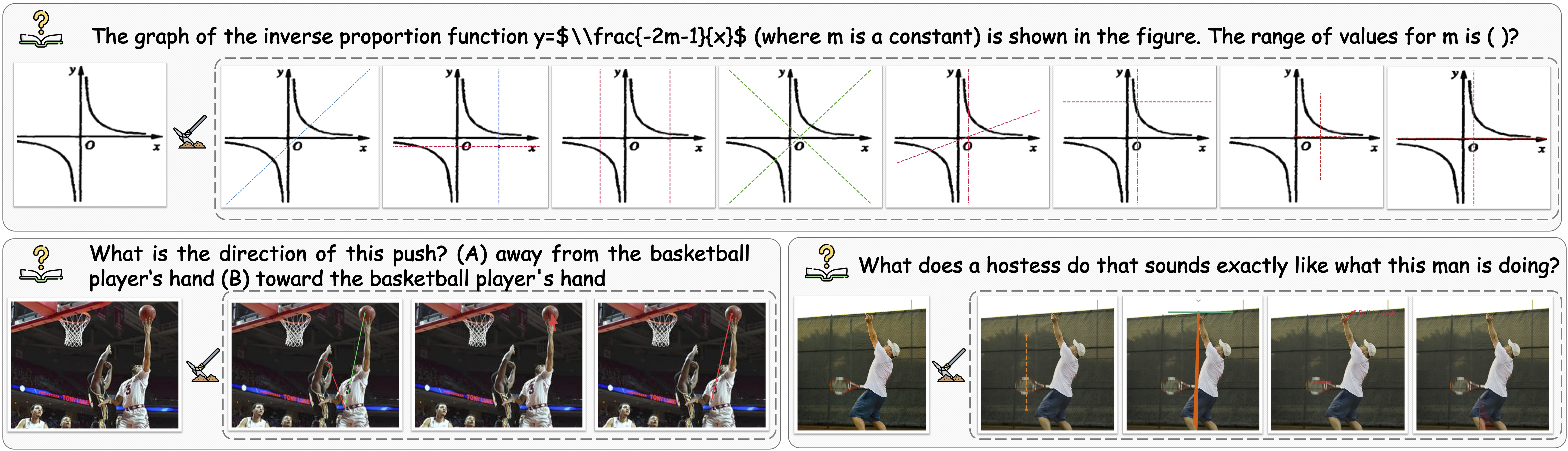}
    }
    \caption{Visualization of a series of samples in rollout sampling.}

    \label{fig:rollout}
\end{figure*}

\section{Experiments}
\label{Experiments}

\subsection{Experimental Setup}

\paragraph{Datasets.}  
\label{sec:dataset}

In this paper, we use two constructed datasets for supervised fine-tuning: \textit{V-Perception-40K} for perceptual alignment and \textit{V-Interaction-400K} for interactive alignment. In the reinforcement learning stage, 40K samples are sampled from We-Math 2.0~\citep{qiao2025we}, MMK12~\citep{MM-Eureka}, ThinkLite~\citep{wang2025sota}, and \textit{V-Interaction-400K}. All datasets are curated in compliance with copyright and licensing regulations. Experiments are conducted on VTBench and three standard visual reasoning benchmarks: MathVision~\citep{wang2024measuring}, We-Math~\citep{qiao2024we}, and VisuLogic~\citep{xu2025visulogic}.

\paragraph{Baselines.}  
We conduct our experiments based on the Qwen2.5-VL-7B~\cite{bai2025qwen2} model and compare our method with two categories of baselines: 
\begin{itemize}[leftmargin=1em, topsep=2pt, itemsep=2pt]
    \item \textbf{Closed-source models:} For example, GPT-4o~\cite{GPT4o}.
    \item \textbf{Open-source general models:} Including models like InternVL3 series~\cite{wang2025internvl3}, LLaVA-OneVision-1.5 series~\citep{an2025llava} and Qwen2.5-VL series~\cite{bai2025qwen2}.
\end{itemize}
Our evaluation is performed using VLMEvalKit~\cite{duan2024vlmevalkit}, with modifications made to the API-calling components due to network constraints. 

\paragraph{Implementation Details.}

For training, we conduct all experiments on $64 \times 8$ H20 GPUs (i.e., 64 nodes with 8 GPUs per node). Both SFT stages use a learning rate of $1 \times 10^{-5}$. The final SFT checkpoint initializes the RL stage, which is trained for one epoch with eight rollouts per iteration, a learning rate of $5 \times 10^{-7}$, and a warm-up ratio of 0.05. 

For VTBench evaluation, we employ two models as judges: Qwen3-235B-A22B and Qwen3-VL-235B-A22B. For Perception Tasks, the judges follow the prompts defined in Table~\ref{tab:lmm-judge-prompt-perception}, while for Instruction-Guided Interaction Tasks, they adopt the prompts specified in Table~\ref{tab:lmm-judge-prompt_IGI}. Given the current limitations in model perception capabilities and the sensitivity of evaluation criteria to prompt variations, we fix the prompt version throughout all experiments to ensure consistent and fair judgments.

For data construction, We employ GPT-5 as the generator $\mathcal{G}$ and adopt Qwen3-VL-232B-A22B as the data checking module $\mathcal{V}$. Furthermore, GPT-4.1 is utilized as the repairer and perform progressive expansion.

\subsection{Main Results}

To thoroughly analyze the effectiveness of our V-Thinker in the field of interactive reasoning, we conduct experimental analyses from both \textit{quantitative} and \textit{qualitative} perspectives:

\paragraph{Quantitative Analysis.} 

As shown in Table~\ref{tab:general_reasoning}, we compare V-Thinker with strong baseline LMMs on VTBench. V-Thinker demonstrates consistent improvements across tasks requiring vision-centric interactions, yielding the following key insights:

\textbf{(1) Perception Challenges Across LMMs.} Despite impressive visual reasoning abilities, perceptual alignment remains a critical bottleneck for advanced models. On VTBench, existing LMMs struggle with fine-grained visual interaction tasks, particularly in identifying spatial relationships and localizing precise points on images. While models like GPT-4o and Qwen2.5-VL excel at visual problem-solving, they underperform on tasks demanding direct visual interaction (e.g., Qwen2.5-VL: 8.8\%). This reveals a substantial gap between general visual reasoning and the specific perceptual grounding required for interactive visual reasoning.

\textbf{(2) Effectiveness of V-Thinker.} Under identical experimental setups, V-Thinker consistently outperforms baseline LMMs across all three interactive reasoning domains, achieving an average accuracy improvement of 12.5\% and maintaining over 6\% gains on individual domains. Notably, it achieves over 22\% performance improvement in the Instruction-Guided Interaction domain. These results underscore V-Thinker's superior interactive thinking capability.

\begin{figure*}[ht!]
    \centering
    \resizebox{\textwidth}{!}{
    \includegraphics{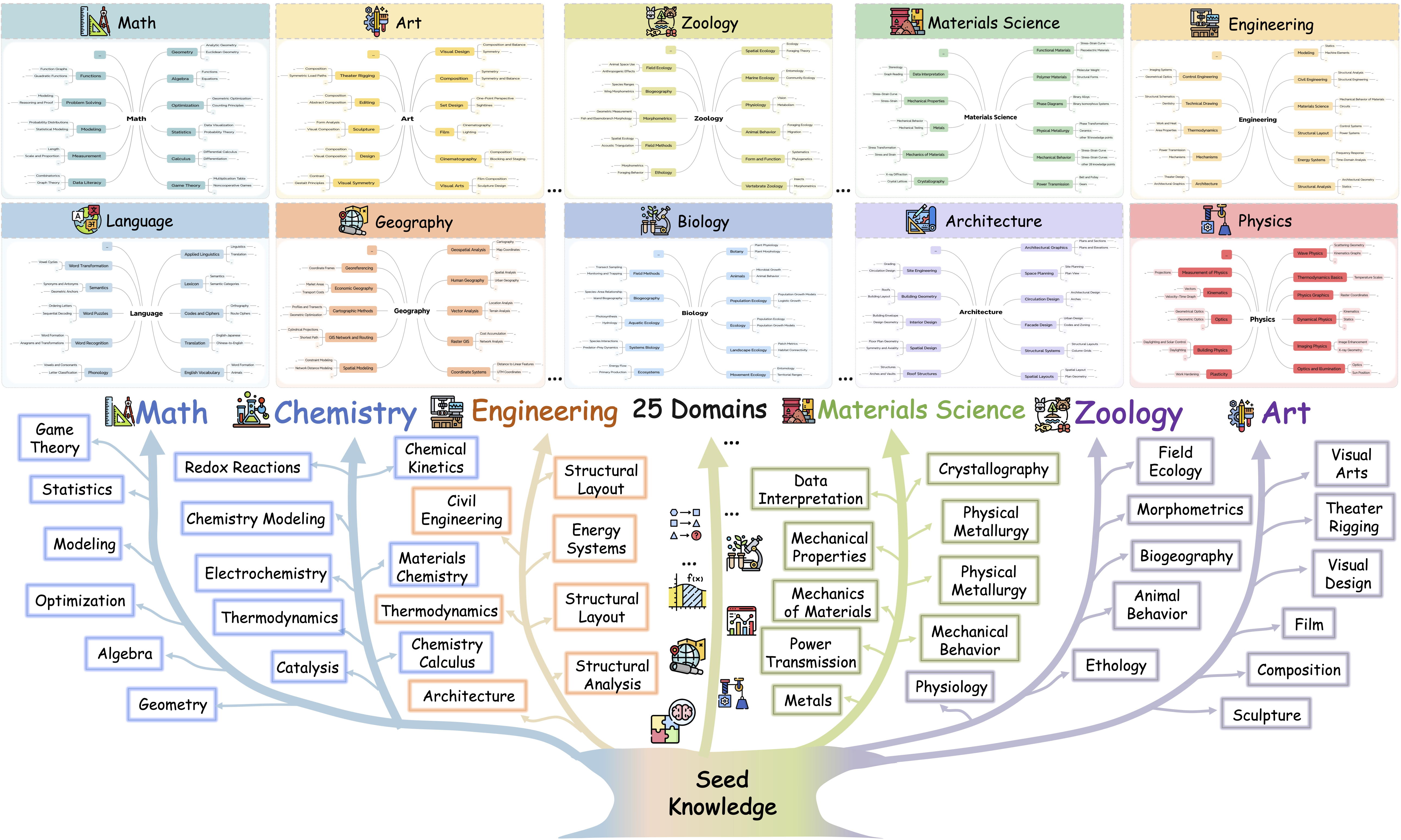}
    }
    \caption{Visualization of the evolved knowledge system through the Data Evolution Flywheel.}

    \label{fig:forest}
\end{figure*}

\begin{figure}[ht!]
    \centering
    \resizebox{0.47\textwidth}{!}{
    \includegraphics{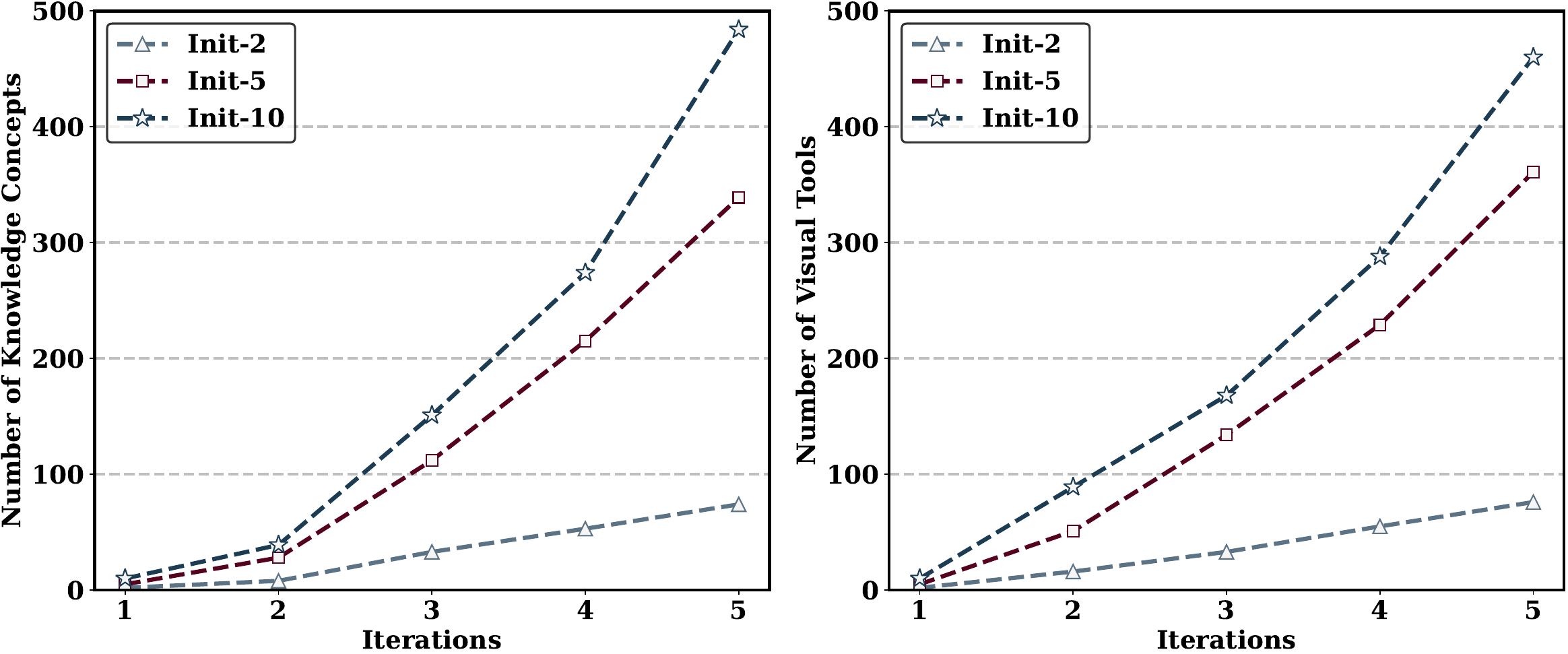}
    }
    \caption{Scaling analysis of the iterations in the Data Evolution Flywheel.}

    \label{fig:zhexian}
\end{figure}

\begin{table}[t!]
\small
\centering
\renewcommand{\arraystretch}{0.9} 

\resizebox{0.9\columnwidth}{!}{%
\begin{tabular}{@{} l ccc ccc @{}}
\toprule
\textbf{Method} & \makecell{\textbf{SFT}\\\textbf{(Per.)}} &
\makecell{\textbf{SFT}\\\textbf{(Int.)}} & \textbf{RL} & \textbf{MVs} & \textbf{WM} & \textbf{VS} \\
\midrule
$M_0$ & \ding{51} & \ding{51} & \ding{51} & 29.3 & 62.8 & 26.6 \\
\midrule
$M_1$ & \ding{51} & \ding{51} & - & 21.4 & 55.5 & 24.8 \\
$M_2$ & - & \ding{51} & \ding{51} & 28.3 & 62.6 & 26.1 \\
\bottomrule
\end{tabular}%
}
\caption{Results of the ablation study. \textbf{MVs}: MathVision; \textbf{WM}: We-Math; \textbf{VS}: VisuLogic.}
\label{tab:ablation}
\end{table}

\begin{figure*}[ht!]
    \centering
    \resizebox{\textwidth}{!}{
    \includegraphics{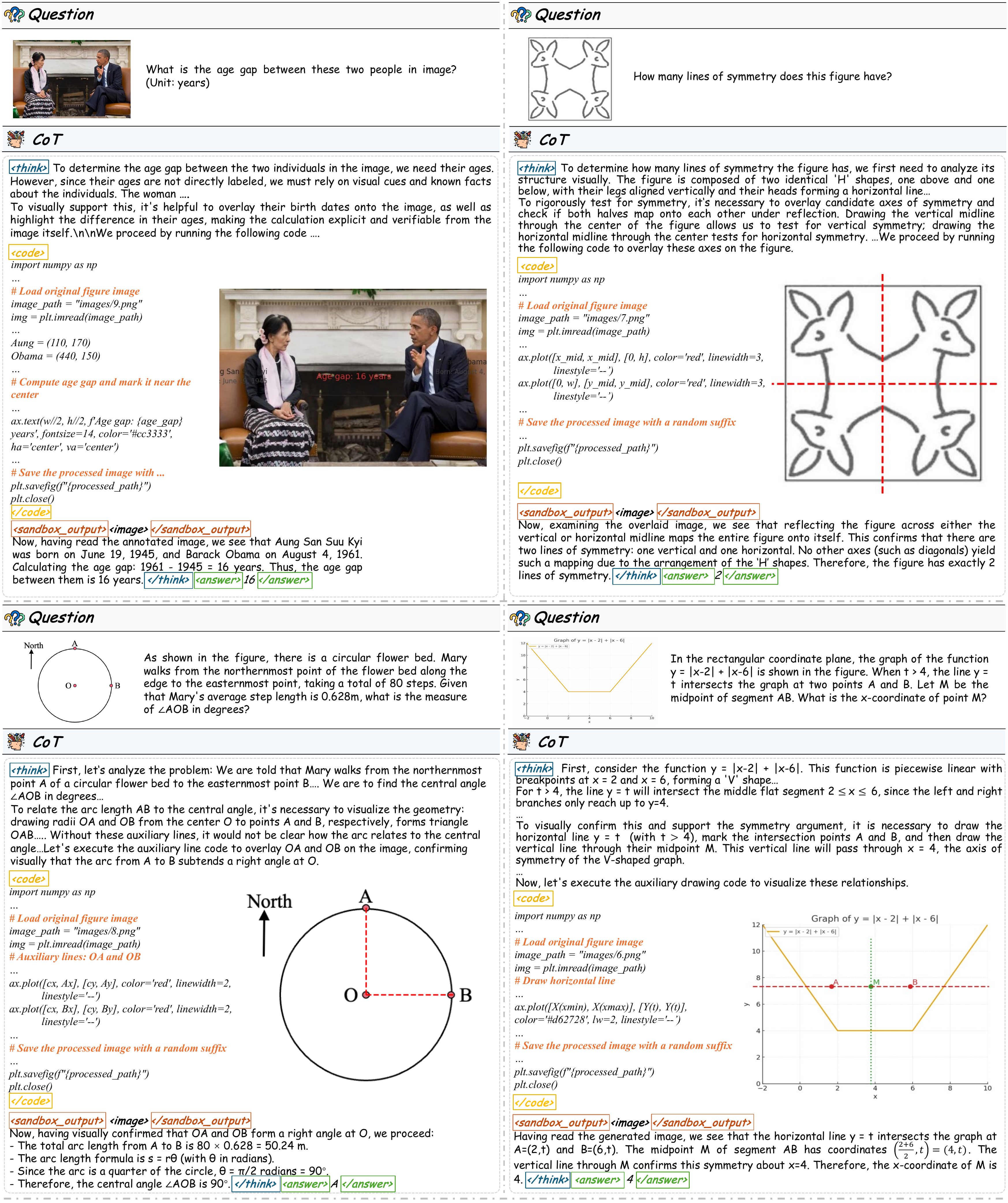}
    }
    \caption{Complete interactive reasoning samples of V-Thinker on open-source benchmarks.}

    \label{fig:cot}
\end{figure*}

\paragraph{Qualitative Analysis.}

To further showcase V-Thinker’s beyond-expected performance in interactive thinking, we analyze three aspects: visual interaction editing, rollout sampling behavior, and fully complete cases.

\textbf{(1) Visual Interaction Editing (Fig.~\ref{fig:caseexp}):} On tasks requiring visual interaction, V-Thinker-7B accurately draws squares in diagrams, fully identifying and labeling instances. For arithmetic tasks that do not strictly require interaction, the model not only produces correct answers but also proactively annotates images to clarify intermediate steps. In real-world scenarios (e.g., meal images, multi-person photos), V-Thinker likewise generates instructive visual annotations, demonstrating stable and fine-grained visual interaction editing.

\textbf{(2) Rollout Sampling Analysis (Fig.~\ref{fig:rollout}):} We quantify sampling results within a single rollout step for the same image. The results show that V-Thinker develops multi-dimensional derivative reasoning and covers a broader solution space during RL stage, reflecting stronger exploration and more diverse reasoning paths.

\textbf{(3) Complete Cases (Fig.~\ref{fig:cot}):} To illustrate the end-to-end interactive thinking process, we present V-Thinker’s visual reasoning trajectory. During reasoning, the model autonomously generates image-editing code and immediately renders edited outputs, externalizing intermediate states and simplifying the reasoning chain—forming a think–edit loop.

Overall, these three lines of evidence jointly demonstrate that V-Thinker exhibits superior and transferable interactive reasoning capabilities.

\subsection{Results on Generalized Reasoning Domain}
To further validate V-Thinker's generalization capability on general reasoning tasks, Table~\ref{tab:general_reasoning} presents its performance across three widely-used benchmarks. V-Thinker-7B demonstrates consistent improvements across three benchmarks, with stable gains in vision-centric mathematics and logical reasoning. Notably, without specific in-domain data introducing, V-Thinker achieves substantial improvements on complex multi-step reasoning tasks such as MathVision (+6.3\%). These results validate that V-Thinker's interactive reasoning paradigm generalizes effectively across diverse reasoning domains.

\subsection{Ablation Study.} To explore the roles of different modules in V-Thinker, we perform an ablation study in general reasoning scenarios, as shown in Table~\ref{tab:ablation}. We have the following observations: \textbf{(1)} The performance of V-Thinker declines when any training phrase is removed, indicating that all progressive curriculum training stages are effective. \textbf{(2)} Removing perception alignment results in notable performance drop, demonstrating that perception capability serves as a crucial foundation for subsequent interactive reasoning. \textbf{(3)} Ablating RL training brings substantial performance degradation, particularly showing over 6\% performance drop in Mathvision and We-Math. This indicates that RL stage is essential for exploring interactive reasoning patterns and enabling effective generalization to general reasoning scenarios.

\subsection{Analysis of the Data Evolution Flywheel}

\paragraph{Data Evolution Flywheel Effectively Expands Knowledge Systems.}  
Figure~\ref{fig:forest} illustrates an evolved knowledge system derived from seed concepts through the Data Evolution Flywheel, encompassing 25 distinct domains. The resulting hierarchical structure reaches a maximum depth of 7 layers and 24,767 nodes, demonstrating the flywheel’s capability to construct diverse, large-scale, and multi-level knowledge representations. This scalable architecture establishes a systematic foundation for domain-specific and cross-domain reasoning, facilitating knowledge organization and synthesis across interactive reasoning tasks.

\paragraph{Scaling Analysis of Evolution Iterations.} To investigate the relationship between evolution iterations and the expansion of knowledge systems and visual tools, Figure~\ref{fig:zhexian} quantifies the incremental growth of knowledge concepts and visual tools derived from synthesized data at each iteration.

We observe two key findings: \textbf{(1) Non-linear Growth with Evolve knowledge Expansion.} As evolution iterations increase, both knowledge concepts and visual tools exhibit non-linear growth without saturation. After five iterations, the system scales to approximately 50$\times$ the initial seed size. This validates the superiority of our Data Evolution Flywheel design, which fundamentally enhances data diversity by discovering orthogonal knowledge concepts and visual tools as anchors, thereby sustaining continuous flywheel momentum. \textbf{(2) Impact of Initial Seed Diversity.} Richer initial knowledge concepts or tool sets yield superior evolution trajectories, underscoring the critical importance of providing diverse seed knowledge concepts as foundational anchors.

\section{Conclusion}

In this paper, we propose V-Thinker, a general-purpose multimodal reasoning assistant that enables interactive, vision-centric thinking through end-to-end reinforcement learning. Our main contributions include: (1) formalizing \textit{Interactive Thinking with Images} and developing an end-to-end framework that bridges visual grounding with code-driven interactive reasoning; (2) proposing a \textbf{Data Evolution Flywheel} that automatically synthesizes, evolves, and verifies datasets across diversity, quality, and difficulty dimensions; (3) introducing a \textbf{Visual Progressive Training Curriculum} that aligns perception and interactive reasoning through a two-stage training framework; and (4) releasing \textbf{VTBench}, an expert-verified benchmark for comprehensive evaluation. Extensive experiments show that V-Thinker consistently outperforms mainstream vision-language baselines, advancing the field of interactive visual reasoning and providing practical insights for future multimodal system development.



{
    \small
    \bibliographystyle{ieeenat_fullname}
    \bibliography{main}
}

\clearpage

\appendix



\twocolumn[
\begin{center}
{\Large \textbf{Appendix}}
\end{center}
]

\tableofcontents

\section{Details of V-Thinker}
\label{sec:supple vthinker}

\subsection{Prompts Used in V-Thinker}






\paragraph{Prompt Templates for Knowledge-Driven and Tool-Driven Synthesis.}

To operationalize the two synthesis pathways in our co-evolution framework, we design two prompt specifications that explicitly encode how new instances are generated under knowledge-driven or tool-driven conditioning.

As shown in Table~\ref{tab:knowledge-driven-prompt}, the knowledge-driven prompt formalizes how sampled knowledge combinations guide the construction of problems, the rendering of figures, and the generation of reasoning trajectories involving visual tools. The underlying knowledge inventory is initialized from the 1{,}819 fundamental principles in We-Math 2.0~\citep{qiao2025we}, and we instantiate both single-concept inputs and compositional combinations, so that the synthesized data can cover a broad range of reasoning objectives.

As shown in Table~\ref{tab:tool-driven-prompt}, the tool-driven prompt specifies the complementary mechanism in which visual tools act as the generative anchor. The initial visual tool system contains 61 visual tools, constructed by abstracting common operations from existing tool-augmented multimodal systems~\citep{li2025mathopeval,wei2025geoint,wu2025vtool,zhou2025perception,zhang2025deepsketcher,zhou2025reinforced,li2025look,zhao2025pyvision,zhang2025thyme}. It defines how tool combinations constrain the image, the tool operations, and the resulting reasoning process, thereby enabling tool-conditioned data construction.

Together, these two prompt templates characterize how knowledge and tools drive generation along orthogonal dimensions. To ensure consistency across all visual tools, the rendering protocol enforces a unified set of conventions: all figures and tool overlays use absolute pixel coordinates, a common top-left origin, and full-bleed canvases without margins. These choices guarantee stable visual structure, precise localization, and reproducible execution across iterations. Further implementation details are provided in the corresponding prompt specifications.

\paragraph{Prompts for Coordinated Calibration}

To support the coordinated calibration stage, we provide three checker prompt templates that separately evaluate (1) answer correctness, (2) integrity of the rendered image, and (3) consistency of auxiliary visual states. These templates are shown in Tables~\ref{tab:checker-answer}, \ref{tab:checker-image}, and \ref{tab:checker-aux}.

For cases where only the textual answer is inconsistent while the visual components remain valid, a dedicated repair prompt (Table~\ref{tab:repair}) reconstructs the question from the visual states to realign the reasoning. Together, these templates form the basis of the calibration mechanism used to refine the initial dataset into a coherent and verified collection.

\paragraph{Prompts for Progressive Expansion}

Progressive expansion is operationalized through two coordinated prompt templates.
The parallel extension prompt (Table~\ref{tab:parallel-extension-prompt}) specifies how additional visual constructions are incorporated to introduce complementary observations beyond the original configuration.
The sequential extension prompt (Table~\ref{tab:sequential-extension-prompt}) prescribes constructions that are conditioned on existing entities or intermediate results, enabling reasoning to unfold through successive stages. These templates define the procedural rules for generating extended visual states and deeper reasoning trajectories, supporting the formation of higher-difficulty instances.

\begin{table*}[t]
\centering
\setlength\tabcolsep{17.5pt}
\renewcommand{\arraystretch}{1.0}
\begin{tabular}{
    p{3.5cm} 
    c c c c
}
\toprule
\multirow{2}{*}{\textbf{Method}}
  & \multicolumn{4}{c}{\textbf{VTBench}} \\
\cmidrule(lr){2-5}
    & \textit{Perception} 
    & \makecell{\textit{Instruct.}\\\textit{Interaction}}
    & \makecell{\textit{Interactive}\\\textit{Reasoning}}
    & \textit{Avg.} \\
\midrule

\rowcolor[RGB]{245,245,250} 
GPT-4o 
& 12.6 & 26.0 & 36.4 & 25.0 \\

Qwen2.5-VL-72B 
& 38.0 & 34.2 & 51.4 & 41.2 \\

\rowcolor[RGB]{245,245,250}
InternVL3-78B 
& 13.8 & 19.0 & 43.4 & 25.4 \\

InternVL3-8B 
& 10.4 & 6.8 & 33.8 & 17.0 \\

\rowcolor[RGB]{245,245,250}
LLaVA-OV-1.5-8B 
& 12.2 & 12.2 & 30.2 & 18.2 \\

InternVL3-2B 
& 3.0 & 3.4 & 22.0 & 9.5 \\

\rowcolor[RGB]{245,245,250}
Qwen2.5-VL-3B
& 2.6 & 2.2 & 28.6 & 11.1 \\

LLaVA-OV-1.5-4B 
& 8.2 & 10.8 & 30.2 & 16.4 \\

\rowcolor[RGB]{245,245,250}
Qwen2.5-VL-7B 
& 12.6 & 8.8 & 31.8 & 17.7 \\

\hdashline

\rowcolor{gray!8}
\textbf{V-Thinker-7B}
& \textbf{18.6} & \textbf{31.6} & \textbf{40.4} & \textbf{30.2} \\

\rowcolor{gray!8}
$\Delta$ (\textit{vs} Qwen2.5-VL-7B)
& \textbf{\textit{\textcolor{red!50!black}{+6.0}}}
& \textbf{\textit{\textcolor{red!50!black}{+22.8}}}
& \textbf{\textit{\textcolor{red!50!black}{+8.6}}}
& \textbf{\textit{\textcolor{red!50!black}{+12.5}}} \\

\bottomrule
\end{tabular}
\caption{Overall performance on VTBench. 
(\textit{Instruct.~Interaction} denotes \textit{Instruction-Guided Interaction}.)}
\label{tab:vtbench_performance}
\end{table*}

\paragraph{Prompts for Perception Data Synthesis}

Perception-oriented data generation is supported by four prompt templates, corresponding to code construction and three levels of perceptual QA.
The figure-construction prompt (Table~\ref{tab:figure-construction-prompt}) governs the generation of executable drawing code and the associated element-level annotations, ensuring that visual content is specified in absolute pixel coordinates and adheres to the structural constraints defined by element relations, element count, and sampled knowledge concepts.

Based on these annotations, three QA prompts convert visual structures into perception tasks along the hierarchical dimensions introduced above.
The surface-level prompt (Table~\ref{tab:surfaceqa}) produces localization questions grounded in explicit element coordinates.
The semantic-level prompt (Table~\ref{tab:semanticqa}) builds questions around structurally defined feature points, requiring spatial interpretation beyond raw coordinates.
The integrated reasoning prompt (Table~\ref{tab:Integratedqa}) constructs tasks that combine perceptual identification with computational reasoning, such as determining derived geometric or structural quantities.

Together, these templates operationalize the construction of $\mathcal{D}_{\text{perception}}$ by transforming synthesized visual structures into systematically layered perceptual QA pairs.




\section{Additional Results and Analyses}

\subsection{Extended Quantitative Analysis}

Table~\ref{tab:vtbench_performance} provides an extended evaluation on VTBench by incorporating additional variants from the Qwen2.5-VL, InternVL3, and LLaVA-OneVision-1.5 families. Across this expanded comparison, two observations remain consistent.

\paragraph{Fine-grained perceptual grounding is consistently weak across model scales.}
Large models such as GPT-4o, Qwen2.5-VL-72B, and InternVL3-78B exhibit limited accuracy on perception-oriented tasks, which reflects the difficulty posed by tasks that require local spatial grounding, such as identifying specific points, intersections, or geometric primitives.

\paragraph{V-Thinker shows consistent improvements in perception and interaction.}
Relative to 7B-scale open-source models, V-Thinker achieves higher accuracy (+6.0\%) on perception-oriented questions and a larger gain (+22.8\%) on instruction-guided interaction, together with stable improvements in visual reasoning (+8.6\%). These trends are consistently observed across the expanded baseline set.

\subsection{Additional Analysis of the Data Evolution Flywheel}

\paragraph{Expanded Knowledge Structures.}
Figure~\ref{fig:zhexian} provides a comprehensive visualization of the knowledge system produced through the Data Evolution Flywheel. Beyond the summary shown in the main paper, the extended structure highlights the breadth and depth achieved through iterative evolution. The resulting hierarchy spans 25 domains and reaches up to 7 layers, forming a graph with 24{,}767 nodes. This expanded view further illustrates how the flywheel progressively enriches conceptual coverage, organizes related concepts into coherent clusters, and builds multi-level structures that support both fine-grained and cross-domain reasoning.

\paragraph{Tool System Analysis.}
Within the Data Evolution Flywheel, the visual tool system co-evolves alongside the knowledge concepts, and the resulting tool set undergoes an additional consolidation stage. Since visual tools are instantiated through executable Python drawing routines, functionally identical operations may differ only in minor parameter settings (for example, line styles or rendering options), which can artificially inflate the apparent diversity of tools without reflecting meaningful distinctions in visual operations. To obtain a more faithful representation of the underlying tool space, we apply an additional round of BGE-based hierarchical clustering to the co-evolved tool set, using a strict similarity threshold of 0.3. The resulting clusters are then normalized through LMM-assisted (GPT-4.1) unification of tool names, followed by manual adjustment to ensure semantic consistency. This refinement phase collapses redundant variants and yields a compact, semantically coherent library of 234 visual tools, providing a clearer and more accurate characterization of the evolved visual tool system.

\section{Details of VTBench}
\label{sec:supple vtbench}

\subsection{Evaluation Dimensions}

VTBench assesses interactive reasoning across three hierarchical dimensions, modeling the progression from perception to adaptive interaction:
\begin{equation*}
\resizebox{0.48\textwidth}{!}{$\textit{Perception} \rightarrow \textit{Instruction-Guided Interaction} \rightarrow \textit{Interactive Reasoning}$}
\end{equation*}

\begin{itemize}[leftmargin=1em, topsep=1pt, itemsep=1pt]
\item \textbf{Perception:} Evaluates fine-grained visual perception, such as identifying or locating specific coordinates.
\item \textbf{Instruction-Guided Interaction:} Tests the ability to execute explicit visual instructions (e.g., drawing lines, labeling regions).
\item \textbf{Interactive Reasoning:} Evaluates reasoning tasks that require visual interaction.
\end{itemize}

\subsection{Data Statistics and Evaluation Metrics}

VTBench contains 1,500 QA pairs (500 per task type) across 9 open-source benchmarks (MathVista~\citep{lu2023mathvista}, MathVision~\citep{wang2024measuring}, MathVerse~\citep{zhang2024mathverse}, Dynamath~\citep{zou2024dynamath}, We-Math~\citep{qiao2024we}, LogicVista~\citep{xiao2024logicvista}, CMM-Math~\citep{liu2024cmm}, CharXiv~\citep{wang2024charxiv} and ZeroBench~\citep{roberts2025zerobench}) covering four domains: \textit{Logical Reasoning, Geometry, Algebra, and Statistics}. Evaluation is conducted under three task-specific metrics:

\begin{itemize}[leftmargin=1em, topsep=2pt, itemsep=2pt]
\item \textbf{Perception Tasks:} Models generate Python code to mark perceived coordinates. The resulting image is compared with annotations by large multimodal models (LMMs).
\item \textbf{Instruction-Guided Interaction Tasks:} Models generate Python code to perform instructed actions. The visual output is compared with expert annotations using LMM judgment.
\item \textbf{Interactive Reasoning Tasks:} Models output final answers, which are evaluated by LLMs for correctness.
\end{itemize}

\section{Detailed Experimental Setup}



\subsection{Implementation Details (Evaluation)}

For VTBench evaluation, we employ two models as judges: Qwen3-235B-A22B and Qwen3-VL-235B-A22B. Their results differ by less than 2\% from GPT-4.1, validating their reliability as open-source evaluators under a unified evaluation protocol. For all other benchmarks used in this work, including MathVision, VisuLogic, and We-Math, we follow the official evaluation procedures specified by each benchmark, including the provided scoring scripts and answer-matching rules.

\subsection{Details of Baselines.}

We evaluate all models on MathVision, VisuLogic, We-Math and VTBench, following each benchmark’s official protocol and reporting accuracy as the primary metric. Our baseline comparison includes a broad spectrum of multimodal reasoning systems. GPT-4o is OpenAI’s flagship multimodal model designed for unified vision–language understanding with strong cross-domain reasoning capability~\citep{GPT4o}. The Qwen2.5-VL family (72B/7B/3B) is an open-source vision–language series that emphasizes visual–language alignment and multimodal reasoning~\citep{bai2025qwen2}. The InternVL3 series (78B/8B/2B) represents an open-source vision–language architecture with enhanced multimodal perception and reasoning capability~\citep{zhu2025internvl3}. The LLaVA-OneVision-1.5 family (8B/4B) consists of open multimodal models incorporating a unified vision encoder and language backbone, optimized for efficient and competitive vision–language reasoning~\citep{an2025llava}. Together, these baselines span a wide range of architectures and parameter scales, enabling a comprehensive and balanced comparison against our model.

\section{Broaden Impact}
\label{sec:supple broaden impact}

\paragraph{Advancing interactive, vision-centered reasoning in multimodal systems.}

Our work encourages a transition from passive visual perception to interactive reasoning within images. Instead of only understanding visual content, V-Thinker can actively engage with visual elements through pointing, annotating, and manipulating structured regions. This shift supports explicit intermediate reasoning steps and improves transparency in visual-language decision making. Such interactive capabilities may influence a wide range of tasks including diagram understanding, scientific figure analysis, embodied perception, and human–AI collaborative interfaces, where traceable and verifiable reasoning is essential.

\paragraph{Enabling a creator-oriented synthetic data paradigm that expands future possibilities.}

V-Thinker revisits the long-standing paradigm in which models act purely as solvers when synthesizing data. This solver-centric view restricts the diversity and structural richness of generated samples, particularly for tasks requiring precise spatial or logical alignment. Our work reveals that modern multimodal models can instead function as creators capable of generating complex visual problems, programmatically rendered images, auxiliary diagrams, and coherent reasoning paths. When integrated with knowledge-driven representations, this creator-oriented paradigm significantly enlarges the design space for synthetic reasoning data. It reduces reliance on handcrafted seeds, enables scalable and controlled curriculum evolution, and offers a foundation for autonomous dataset construction and simulation-based training. This paradigm shift may inspire future research on model-driven data ecosystems, controllable synthetic corpora, and new forms of interaction-centric supervision for robust multimodal intelligence.

\paragraph{Bridging foundational models with practical, tool-oriented applications.}

The structured visual interactions and interpretable reasoning processes supported by V-Thinker help narrow the gap between multimodal foundation models and their real-world deployment. The ability to generate, manipulate, and reason with structured visual information can benefit scientific analysis, education technologies, robotics perception, and interactive decision-support systems. The explicit reasoning traces also promote safer and more accountable AI behavior, which is crucial in settings that demand correctness and transparency. By supporting precise visual alignment and interpretable reasoning, our work contributes to building reliable multimodal systems that operate effectively in complex visual environments.

\section{Limitation}

V-Thinker explores a generalized paradigm for interactive visual reasoning, where models perform reasoning through visual interactions. However, it still has several limitations. Due to computational constraints, our current model iteration remains limited in scale and exhibits reduced capability in knowledge-intensive or domain-specific tasks.
Our primary goal in this work is to establish a unified framework for interactive reasoning.
In future work, we will further optimize the model and enhance its robustness and generalization across diverse reasoning domains.
Moreover, through V-Thinker, we believe that as model capabilities continue to advance, both the paradigm of data construction and the upper bound of model reasoning should be re-examined and redefined. We envision future models achieving increasingly natural and human-like forms of visual interaction and reasoning.

\clearpage

\begin{table*}[t]
  \vspace{-0.5cm}
  \noindent
  \begin{tcolorbox}[
    width=\textwidth,
    before=\noindent,
    fontupper=\tiny,
    colback=gray!2!white,       
    colframe=gray!45!black,     
    colbacktitle=gray!15!white, 
    coltitle=black,             
    fonttitle=\bfseries,
    title={LMM Judge Prompt (Perception Task)}
  ]
You are an expert visual evaluator. Your goal is to determine whether the point(s) marked in the [Generated Image] meaningfully implement the visual operation described in the [Instruction], using the [Ground Truth Image] as the semantic reference.\\[2pt]
Your evaluation should prioritize conceptual correctness while allowing moderate spatial deviation.\\[4pt]
---\\[2pt]
\textbf{Judging Criteria}\\[4pt]
Judgement = 1 (Consistent)\\
Output \textbf{1} if ALL of the following are satisfied:\\[4pt]

\textbf{1. Correct Visual Concept}
\begin{itemize}
    \setlength\itemsep{0em}
    \renewcommand{\labelitemi}{-} 
    \item The marked point(s) reflect the correct visual idea from the instruction.
    \item The point(s) are placed on the correct target feature or structure (for example, the intended corner, midpoint, center, or intersection), even if the position is not exact.\\[4pt]
\end{itemize}

\textbf{2. Minor Spatial Deviation Allowed}
\begin{itemize}
    \setlength\itemsep{0em}
    \renewcommand{\labelitemi}{-}          
    \renewcommand{\labelitemii}{\textbullet} 
    \item The modification may deviate moderately, but only if:
    \begin{itemize}
        \item It remains clearly associated with the intended feature,
        \item The positional error is small enough that the point still unambiguously indicates the correct target,
        \item The structural relation remains recognizable (e.g., the point lies on or near the correct line segment, sits close to the correct vertex, or falls within a reasonable neighborhood of the correct intersection).
    \end{itemize}
    \item Examples of acceptable deviation:
    \begin{itemize}
        \item A midpoint marker slightly off-center but still indicating the middle region of the correct segment,
        \item A vertex marker somewhat offset but still within the local vicinity of the intended vertex.\\[4pt]
    \end{itemize}
\end{itemize}

\textbf{3. Stylistic Variations Ignored}
\begin{itemize}
    \setlength\itemsep{0em}
    \renewcommand{\labelitemi}{-}
    \item Differences in point color, size, shape, or style (for example, dot vs small circle vs ``x'') must be ignored.
    \item Minor rendering artifacts that do not change the intended target should also be ignored.
\end{itemize}

\vspace{4pt}
Judgement = 0 (Inconsistent)\\
Output \textbf{0} if ANY of the following hold:\\[2pt]

\textbf{1. Wrong Concept}
\begin{itemize}
    \setlength\itemsep{0em}
    \renewcommand{\labelitemi}{-}
    \item The point(s) are placed on the wrong feature or structure (for example, a different vertex, a different segment, a different circle, or an unrelated location).
    \item The marking does not correspond to the operation described in the instruction.\\[2pt]
\end{itemize}

\textbf{2. Major Conceptual Misalignment}
\begin{itemize}
    \setlength\itemsep{0em}
    \renewcommand{\labelitemi}{-}
    \item The location of the point(s) is so far from the intended target that the underlying operation is no longer recognizable, even under generous tolerance.\\[2pt]
\end{itemize}

\textbf{3. Missing or Insufficient Marking}
\begin{itemize}
    \setlength\itemsep{0em}
    \renewcommand{\labelitemi}{-}
    \item The required point(s) are missing, or the markings are too incomplete or ambiguous to reflect the instruction.\\[2pt]
\end{itemize}

\textbf{4. No Effective Change}
\begin{itemize}
    \setlength\itemsep{0em}
    \renewcommand{\labelitemi}{-}
    \item The [Generated Image] is effectively identical to the [Original Image].
\end{itemize}

\vspace{2pt}
---\\[2pt]
\textbf{Output Format}
\begin{itemize}
    \setlength\itemsep{0em}
    \renewcommand{\labelitemi}{-}
    \item If consistent, output 1.
    \item If inconsistent, output 0.
\end{itemize}
Output ONLY 0 or 1. Do not provide any explanation.\\[4pt]
---\\[2pt]
\textbf{Evaluation Inputs}\\[4pt]
[Original Image] <|vision\_start|><|image\_pad|><|vision\_end|>\\[2pt]
[Generated Image] <|vision\_start|><|image\_pad|><|vision\_end|>\\[2pt]
[Ground Truth Image] <|vision\_start|><|image\_pad|><|vision\_end|>\\[4pt]
[Instruction]: \{instruction\}\\[4pt]
Provide your judgement.\\
Judgement:
  \end{tcolorbox}
  \caption{Prompts for LMM Judge Prompt (Perception Task).}
  \label{tab:lmm-judge-prompt-perception}
\end{table*}

\begin{table*}[t]
  \vspace{-0.5cm}
  \noindent
  \begin{tcolorbox}[
    width=\textwidth,
    before=\noindent,
    fontupper=\tiny,
    colback=gray!2!white,       
    colframe=gray!45!black,     
    colbacktitle=gray!15!white, 
    coltitle=black,             
    fonttitle=\bfseries,
    title={LMM Judge Prompt (Instruction-Guided Interaction Task)}
  ]
You are an expert visual evaluator. Your goal is to determine whether the modifications in the [Generated Image] meaningfully implement the visual operation described in the [Instruction], using the [Ground Truth Image] as the semantic reference.\\[2pt]

Your evaluation should prioritize conceptual correctness while allowing moderate spatial deviation.\\[4pt]
---\\[2pt]
\textbf{Judging Criteria}\\[4pt]
Judgement = 1 (Consistent)\\
Output \textbf{1} if ALL of the following are satisfied:\\[4pt]

\textbf{1. Correct Visual Concept}
\begin{itemize}
    \setlength\itemsep{0em}
    \renewcommand{\labelitemi}{-} 
    \renewcommand{\labelitemii}{\textbullet}          
    \item The modification reflects the correct visual idea from the instruction.
    \item The modification targets the correct general location or structure, even if not precise.\\[4pt]
\end{itemize}

\textbf{2. Broader Tolerance for Spatial Deviation}
\begin{itemize}
    \setlength\itemsep{0em}
    \renewcommand{\labelitemi}{-} 
    \renewcommand{\labelitemii}{\textbullet}    
    \item The modification may deviate significantly, as long as:
    \begin{itemize}
        \item it is within the correct overall area,
        \item it connects or marks the correct conceptual components,
        \item and the intended structural relation remains recognizable.
    \end{itemize}
    \item Examples of acceptable deviation:
    \begin{itemize}
        \item endpoints not exactly touching but pointing to correct vertices,
        \item a region roughly outlined but not tightly aligned,
        \item a line slightly tilted or offset but indicating the right relation.\\[4pt]
    \end{itemize}
\end{itemize}

\textbf{3. Stylistic Variations Ignored}
\begin{itemize}
    \setlength\itemsep{0em}
    \renewcommand{\labelitemi}{-}
    \item Differences in color, stroke thickness, line style, opacity, and rendering artifacts must be ignored.
\end{itemize}

\vspace{4pt}
Judgement = 0 (Inconsistent)\\
Output \textbf{0} if ANY of the following hold:\\[2pt]

\textbf{1. Wrong Concept}
\begin{itemize}
    \setlength\itemsep{0em}
    \renewcommand{\labelitemi}{-}
    \item The modification represents the wrong type of operation (e.g., a line instead of a point, marking the wrong region).
    \item The wrong endpoints, wrong angle, or wrong region are used.\\[2pt]
\end{itemize}

\textbf{2. Major Conceptual Misalignment}
\begin{itemize}
    \setlength\itemsep{0em}
    \renewcommand{\labelitemi}{-}
    \item The modification is placed in a way that the intended structure is no longer recognizable, even with generous tolerance.\\[2pt]
\end{itemize}

\textbf{3. Missing or Insufficient Modification}
\begin{itemize}
    \setlength\itemsep{0em}
    \renewcommand{\labelitemi}{-}
    \item The required visual change is absent or too incomplete to reflect the instruction.\\[2pt]
\end{itemize}

\textbf{4. No Effective Change}
\begin{itemize}
    \setlength\itemsep{0em}
    \renewcommand{\labelitemi}{-}
    \item The [Generated Image] is effectively identical to the [Original Image].
\end{itemize}

\vspace{2pt}
---\\[2pt]
\textbf{Output Format}
\begin{itemize}
    \setlength\itemsep{0em}
    \renewcommand{\labelitemi}{-}
    \item If consistent, output 1.
    \item If inconsistent, output 0.
\end{itemize}
Output ONLY 0 or 1. Do not provide any explanation.\\[4pt]
---\\[2pt]
\textbf{Evaluation Inputs}\\[4pt]
[Original Image] <|vision\_start|><|image\_pad|><|vision\_end|>\\[2pt]
[Generated Image] <|vision\_start|><|image\_pad|><|vision\_end|>\\[2pt]
[Ground Truth Image] <|vision\_start|><|image\_pad|><|vision\_end|>\\[4pt]
[Instruction]: \{instruction\}\\[4pt]
Provide your judgement.\\
Judgement:
  \end{tcolorbox}
  \caption{Prompts for LMM Judge Prompt (Instruction-Guided Interaction Task).}
  \label{tab:lmm-judge-prompt_IGI}
\end{table*}

\begin{table*}[t]
  \vspace{-0.5cm}
  \noindent
  \begin{tcolorbox}[                 
    width=\textwidth,        
    before=\noindent,       
    fontupper=\tiny, 
    colback=gray!2!white,          
    colframe=gray!45!black,        
    colbacktitle=gray!15!white,    
    coltitle=black,                
    fonttitle=\bfseries,
    title={Generator (Knowledge-driven)}
  ]
You are an expert exam question designer and visualization engineer. I will provide you with a knowledge point and a category. Please generate three high-quality exam questions that meet the following requirements:\\
    \quad • Quantity: output exactly three questions\\
    \quad • Question type: multiple choice or fill-in-the-blank\\
    \quad • Difficulty: one easy, one medium, one hard\\
    \quad • Figure: each problem must require a figure to solve; the figure is generated using pure Python code\\
    \quad • LaTeX: all math expressions must be written in LaTeX\\
    \quad • Aesthetics: the figure must be clear and visually appealing, with proper labels if needed; it can omit a title\\
    \quad • If the figure contains text or symbols, ensure the font and symbols are aesthetically pleasing and consistent\\
    \quad • Ensure the correctness of the provided answer\\
    \quad • Auxiliary lines: the problem must be solvable only after adding one or more auxiliary lines in the figure.\\
    \quad • In the solution, explicitly explain which auxiliary lines are added and why they work.\\
    \quad • The auxiliary lines must be those truly required for solving the problem (not optional or cosmetic).\\
    \quad • A valid auxiliary line should:\\
    \qquad – Be necessary (without it, the problem cannot be solved efficiently or at all),\\
    \qquad – Create useful geometric structures (e.g., similar triangles, right triangles, symmetry, equal radii),\\
    \qquad – Connect meaningful points (such as vertices, midpoints, centers, or intersections),\\
    \qquad – Have a clear role in the reasoning (explainable in the solution).\\
    \quad • Do not include instructions in the problem statement that explicitly tell the student to add auxiliary lines.\\
    \quad • If multiple knowledge points are provided, each generated question must require applying them together.\\
    \quad • In addition to descriptions, also summarize the auxiliary tool construction type in concise English terms (e.g., ``perpendicular line'', ``angle bisector'', ``midpoint connection''). This summary should not influence the figure or solution construction, only be added afterward.\\[4pt]
    ---\\[4pt]
    Python Figure Rules (for ALL original figure code blocks):\\
    \quad • Use only \texttt{matplotlib} and \texttt{numpy}\\
    \quad • In the figure, text and symbols must not use LaTeX rendering in \texttt{matplotlib} (do not enable \texttt{text.usetex}; avoid \texttt{\$...\$}); use standard non-LaTeX fonts\\
    \quad • Code must be self-contained and directly runnable (import, setup, \texttt{plt.show()} at the end)\\
    \quad • Figure size should be appropriate, with readable labels and no unnecessary clutter\\
    \quad • No saving files, printing text, or using external data\\
    \quad • Output must explicitly set canvas width/height and DPI, then create a full-bleed axes with no margins and a top-left origin in pixel space:\\
    \qquad – Define:\\
    \qquad\quad \texttt{w, h = ...}\\
    \qquad\quad \texttt{DPI = ...}\\
    \qquad – Boilerplate (exact order):\\
    \qquad\quad \texttt{fig = plt.figure(figsize=(w/DPI, h/DPI), dpi=DPI)}\\
    \qquad\quad \texttt{ax = fig.add\_axes([0,0,1,1])  \# full-bleed, no borders}\\
    \qquad\quad \texttt{ax.set\_xlim(0, w)}\\
    \qquad\quad \texttt{ax.set\_ylim(h, 0)  \# top-left origin (y downwards)}\\
    \qquad\quad \texttt{ax.axis('off')}\\
    \qquad\quad \texttt{ax.set\_aspect('equal')}\\
    \qquad – End every script with: \texttt{plt.show()}\\
    \quad • Do NOT use \texttt{plt.tight\_layout()}\\
    \quad • Do NOT save files inside the code (no \texttt{fig.savefig}, etc.)\\[4pt]
    ---\\[4pt]
    Auxiliary-lines Code Rules (for ALL auxiliary-line code blocks):\\
    \quad • Load the original figure image ONLY via \texttt{plt.imread(image\_path)}.\\
    \quad • Assume NO prior canvas info (do not reuse \texttt{w}, \texttt{h}, or \texttt{DPI} from the original figure).\\
    \quad • Infer dimensions directly from the loaded image:\\
    \qquad \texttt{img = plt.imread(image\_path)}\\
    \qquad \texttt{h, w = img.shape[:2]}\\
    \qquad \texttt{fig = plt.figure()}\\
    \qquad \texttt{ax = fig.add\_axes([0,0,1,1])}\\
    \qquad \texttt{ax.set\_xlim(0, w)}\\
    \qquad \texttt{ax.set\_ylim(h, 0)  \# top-left origin}\\
    \qquad \texttt{ax.axis('off')}\\
    \qquad \texttt{ax.set\_aspect('equal')}\\
    \qquad \texttt{ax.imshow(img)}\\
    \quad • Oracle-capable construction (preferred for exactness):\\
    \qquad – You may directly reuse or reconstruct the exact geometric parameters/coordinates used in the original figure’s construction (e.g., vertex coordinates, lengths, radii, centers, slopes), as long as auxiliary-line endpoints are computed and drawn in absolute pixel coordinates consistent with the displayed image.\\
    \quad • Perception → Geometry → Overlay (fallback):\\
    \qquad 1. Perceive anchors (deterministically identify landmarks/edges/symmetry/tangency).\\
    \qquad 2. Compute exact pixel endpoints for the auxiliary lines from the perceived anchors.\\
    \qquad 3. Overlay with \texttt{ax.plot()} / \texttt{ax.scatter()} in absolute pixel coordinates.\\
    \qquad – Special case: if perception directly yields absolute coordinates, step (2) can be omitted.\\
    \quad • Do NOT use \texttt{DPI} anywhere in auxiliary-line code.\\
    \quad • No printing, no saving, no external data or libraries.\\
    \quad • End with \texttt{plt.show()}\\[4pt]
    ---\\[4pt]
    Knowledge point: \{knowledge\_point\}\\[2pt]
    Note: Each of the three generated questions must be designed to simultaneously assess the provided knowledge points in combination, not individually.\\[4pt]
    ---\\[4pt]
    Output Format:\\
    Output exactly three JSON objects inside a JSON array (no extra explanations, no backticks). Each object must follow this schema:\\[2pt]
    \texttt{\{\{}\\
    \quad \texttt{"Idx": "\{idx\}",}\\
    \quad \texttt{"Tag": "\{subject\_area\}",}\\
    \quad \texttt{"Knowledge point": "\{knowledge\_point\}",}\\
    \quad \texttt{"difficulty": "easy" \textbackslash{} "medium" \textbackslash{} "hard",}\\
    \quad \texttt{"problem\_type": "multiple choice or fill\_in\_blank",}\\
    \quad \texttt{"python\_code": "Pure Python code as a string, directly runnable (generates the original figure)",}\\
    \quad \texttt{"python\_code\_auxiliary": "Pure Python code as a string, directly runnable (loads the saved figure image via plt.imread and overlays auxiliary lines in pixel coordinates)",}\\
    \quad \texttt{"Question": "Problem statement in Markdown with LaTeX.",}\\
    \quad \texttt{"answer": "string",}\\
    \quad \texttt{"solution\_markdown": "Step-by-step solution in Markdown with LaTeX, where the auxiliary lines are naturally introduced and utilized as part of the reasoning process (CoT).",}\\
    \quad \texttt{"auxiliary\_lines\_description": "Brief description of the auxiliary lines to be added (what and where, in words)",}\\
    \quad \texttt{"auxiliary\_tools\_summary": ["List of concise English action names of auxiliary tool types, e.g., ['Construct Symmetry Axis', 'Construct Tangent Line']]}\\
    \texttt{\}, ...\}}\\
  \end{tcolorbox}
  \caption{Prompts for knowledge-driven evolution.}
  \label{tab:knowledge-driven-prompt}
\end{table*}

\begin{table*}[t]
  \vspace{-0.5cm}
  \noindent
  \begin{tcolorbox}[                 
    width=\textwidth,        
    before=\noindent,       
    fontupper=\tiny, 
    colback=gray!2!white,
    colframe=gray!45!black,
    colbacktitle=gray!15!white,
    coltitle=black,
    fonttitle=\bfseries,
    title={Generator (Tool-driven)}
  ]
You are an expert exam question designer and visualization engineer. I will provide you with a task name and one or more tool names. Please generate three high-quality exam questions that meet the following requirements:\\
(You may be given one or more task names. If you find that the task(s) and tool(s) are difficult to align naturally, you may creatively interpret or extend the task, but you must still follow the required output format exactly.)\\[3pt]
• Quantity: output exactly three questions\\
• Question type: multiple choice or fill-in-the-blank\\
• Difficulty: one easy, one medium, one hard\\
• Figure: each problem must require a figure to solve; the figure is generated using pure Python code\\
• LaTeX: all math expressions must be written in LaTeX\\
• Aesthetics: the figure must be clear and visually appealing, with proper labels if needed; it can omit a title\\
• If the figure contains text or symbols, ensure the font and symbols are aesthetically pleasing and consistent\\
• Ensure the correctness of the provided answer\\
• Auxiliary tools: the problem must be solvable only after adding one or more auxiliary tools in the figure.\\
• The purpose of the exam question is not to test whether the student knows how to use the tool, but to design a problem where the tool is naturally and necessarily required to solve it.\\
• In the solution, explicitly explain which auxiliary tools are added and why they work.\\
• The auxiliary tools must be those truly required for solving the problem (not optional or cosmetic).\\
• A valid auxiliary tool should:\\
\quad – Be necessary (without it, the problem cannot be solved efficiently or at all),\\
\quad – Create useful geometric or algebraic structures (e.g., similar triangles, symmetry, tangent, midpoints),\\
\quad – Connect meaningful points (such as vertices, midpoints, centers, or intersections),\\
\quad – Have a clear role in the reasoning (explainable in the solution).\\
• Do not include instructions in the problem statement that explicitly tell the student to add auxiliary tools.\\
• Additionally, provide a fine-grained hierarchical knowledge point list that captures the layered concepts involved in solving the problem.\\
\quad – Each element must be a full-path string from broad domain to specific property, e.g.:\\
\quad\quad (Geometry)-(Plane Geometry)-(Basic Plane Figures)-(Triangle)-(Properties of Isosceles and Equilateral Triangles)-(Properties of Sides)\\
\quad – If multiple knowledge points are involved, return multiple path strings in a list.\\
\quad – This list should not influence the figure or solution construction, only be added afterward.\\
---\\
\textbf{Python Figure Rules (for ALL original figure code blocks):}\\
• Use only matplotlib and numpy\\
• In the figure, text and symbols must not use LaTeX rendering in matplotlib (do not enable text.usetex; avoid \$…\$); use standard non-LaTeX fonts\\
• Code must be self-contained and directly runnable (import, setup, plt.show() at the end)\\
• Figure size should be appropriate, with readable labels and no unnecessary clutter\\
• No saving files, printing text, or using external data\\
• Output must explicitly set canvas width/height and DPI, then create a full-bleed axes with no margins and a top-left origin in pixel space:\\
\quad – Define: \\
\quad\quad \texttt{w, h =} \\
\quad\quad \texttt{DPI =} \\
\quad – Boilerplate (exact order):\\
\quad\quad \texttt{fig = plt.figure(figsize=(w/DPI, h/DPI), dpi=DPI)}\\
\quad\quad \texttt{ax = fig.add\_axes([0,0,1,1]) \# full-bleed, no borders}\\
\quad\quad \texttt{ax.set\_xlim(0, w)}\\
\quad\quad \texttt{ax.set\_ylim(h, 0) \# top-left origin (y downwards)}\\
\quad\quad \texttt{ax.axis('off')}\\
\quad\quad \texttt{ax.set\_aspect('equal')}\\
\quad – End every script with: \texttt{plt.show()}\\
• Do NOT use \texttt{plt.tight\_layout()}\\
• Do NOT save files inside the code (no \texttt{fig.savefig}, etc.)\\[6pt]
---\\
\textbf{Auxiliary-tools Code Rules (for ALL auxiliary-line code blocks):}\\
• Load the original figure image ONLY via \texttt{plt.imread(image\_path)}.\\
• Assume NO prior canvas info (do not reuse w, h, or DPI from the original figure).\\
• Infer dimensions directly from the loaded image:\\
\quad \texttt{img = plt.imread(image\_path)}\\
\quad \texttt{h, w = img.shape[:2]}\\
\quad \texttt{fig = plt.figure()}\\
\quad \texttt{ax = fig.add\_axes([0,0,1,1])}\\
\quad \texttt{ax.set\_xlim(0, w)}\\
\quad \texttt{ax.set\_ylim(h, 0) \# top-left origin}\\
\quad \texttt{ax.axis('off')}\\
\quad \texttt{ax.set\_aspect('equal')}\\
\quad \texttt{ax.imshow(img)}\\
• Oracle-capable construction (preferred for exactness):\\
\quad – You may directly reuse or reconstruct the exact geometric parameters/coordinates used in the original figure’s construction\\
\quad\quad (e.g., vertex coordinates, lengths, radii, centers, slopes), as long as auxiliary-line endpoints are computed and drawn in absolute pixel coordinates consistent with the displayed image.\\
• Perception → Geometry → Overlay (fallback):\\
\quad 1. Perceive anchors (deterministically identify landmarks/edges/symmetry/tangency).\\
\quad 2. Compute exact pixel endpoints for the auxiliary lines from the perceived anchors.\\
\quad 3. Overlay with \texttt{ax.plot()} / \texttt{ax.scatter()} in absolute pixel coordinates.\\
\quad – Special case: if perception directly yields absolute coordinates, step (2) can be omitted.\\
• Do NOT use DPI anywhere in auxiliary-line code.\\
• No printing, no saving, no external data or libraries.\\
• End with \texttt{plt.show()}\\
---\\
Task: \{task\_name\}\\
Tool: \{tool\_name\}\\
---\\
\textbf{Output Format:}\\
Output exactly three JSON objects inside a JSON array (no extra explanations, no backticks).  
Each object must follow this schema:\\[4pt]
\texttt{\{\{}\\
\quad \texttt{"Idx": "\{idx\}",}\\
\quad \texttt{"Task": "\{task\_name\}",}\\
\quad \texttt{"Tool": "\{tool\_name\}",}\\
\quad \texttt{"difficulty": "easy" \textbackslash{} "medium" \textbackslash{} "hard",}\\
\quad \texttt{"problem\_type": "multiple choice or fill\_in\_blank",}\\
\quad \texttt{"python\_code": "Pure Python code as a string, directly runnable (generates the original figure)",}\\
\quad \texttt{"python\_code\_auxiliary": "Pure Python code as a string, directly runnable (loads the saved figure image via plt.imread and overlays auxiliary tools in pixel coordinates)",}\\
\quad \texttt{"Question": "Problem statement in Markdown with LaTeX.",}\\
\quad \texttt{"answer": "string",}\\
\quad \texttt{"solution\_markdown": "Step-by-step solution in Markdown with LaTeX, where the auxiliary lines are naturally introduced and utilized as part of the reasoning process (CoT).",}\\
\quad \texttt{"auxiliary\_tools\_description": "Brief description of the auxiliary tools to be added (what and where, in words)",}\\
\quad \texttt{"knowledge\_point\_hierarchy": [e.g.}\\
\quad\quad \texttt{"(Geometry)-(Plane Geometry)-(Basic Plane Figures)-(Triangle)-(Properties of Isosceles and Equilateral Triangles)-(Properties of Sides)",}\\
\quad\quad \texttt{"(Geometry)-(Plane Geometry)-(Basic Plane Figures)-(Triangle)-(Properties of Isosceles and Equilateral Triangles)-(Properties of Angles)"]}\\
\texttt{\}, ...\}}\\

  \end{tcolorbox}
  \caption{Prompts for tool-driven generation.}
  \label{tab:tool-driven-prompt}
\end{table*}

\begin{table*}[t]
  \vspace{-0.5cm}
  \noindent
  \begin{tcolorbox}[                 
    width=\textwidth,        
    before=\noindent,       
    fontupper=\footnotesize, 
    colback=gray!2!white,
    colframe=gray!45!black,
    colbacktitle=gray!15!white,
    coltitle=black,
    fonttitle=\bfseries,
    title={Checker (Q\&A)}
  ]
You are an extremely strict and skeptical problem auditor. Your core task is to identify ``solution to be evaluated'' errors.\\[4pt]

\textbf{Core Instructions:}\\
\quad • \textbf{Critically Compare:} Review the ``Provided Solution Process'' and ``Provided Answer''. Your goal is to find all logical fallacies, calculation errors, or missing steps. Do not be misled or influenced by its reasoning process.\\
\quad • \textbf{Scoring:} Based on your independent analysis and comparison, use the following rules to assign a score.\\[4pt]

\textbf{Inputs:}\\
\quad Question: \texttt{"\{question}\}"\\
\quad Provided Answer: \texttt{"\{answer\}"}\\
\quad Primary Image Path: \texttt{"\{image\}"}\\
\quad Solution Process: \texttt{"\{solution\}"}\\[4pt]

\textbf{Scoring (conservative; deduct for ANY issue; ignore drawing issues here):}\\
\quad • Start from 10 ONLY if EVERY step is logically valid, calculations are correct, and the final answer exactly matches (value/units/format).\\
\quad • By using the provided Question, recalculate the problem and check if the answer is completely consistent with Provided Answer. If not, deduct 5 points.\\
\quad • Check if there are any logical problems or reasoning errors in the problem-solving process. If so, deduct 5 points.\\
\quad • Deduct 3--5 for each logical gap, unstated assumption, undefined term, or missing critical step.\\
\quad • Deduct 2--4 for misapplied theorems or arithmetic/algebra mistakes (even minor).\\
\quad • Deduct 4--6 if the final value/units/format does not exactly match the provided answer.\\
\quad • Answer/question type mismatch (e.g., numeric answer for MCQ, option letter for fill-in): set score $\leq 3$.\\
\quad • Clamp to [0,10]; prefer lower scores when uncertain.\\[4pt]

\textbf{Output Requirement:}\\
\quad Output \textbf{ONLY} one integer 0--10.\\
  \end{tcolorbox}
  \caption{Prompt for Q\&A correctness checking.}
  \label{tab:checker-answer}
\end{table*}


\begin{table*}[t]
  \vspace{-0.3cm}
  \noindent
  \begin{tcolorbox}[                 
    width=\textwidth,        
    before=\noindent,       
    fontupper=\footnotesize, 
    colback=gray!2!white,
    colframe=gray!45!black,
    colbacktitle=gray!15!white,
    coltitle=black,
    fonttitle=\bfseries,
    title={Checker (Original Image)}
  ]
You are an IMAGE VALIDITY auditor. Your ONLY task is to evaluate if the shapes in the diagram are drawn completely and correctly.\\[4pt]

\textbf{Important:}\\
\quad • \textbf{IGNORE} the text question and any labels in the image.\\
\quad • Your evaluation must NOT consider semantic accuracy (whether the diagram answers the question).\\
\quad • Focus purely on the drawing quality.\\[4pt]

\textbf{Inputs:}\\
\quad Question: \texttt{"\{question}\}"\\
\quad Primary Image Path: \texttt{"\{image\}"}\\[4pt]

\textbf{Scoring Criteria (Based ONLY on Drawing Quality):}\\[2pt]
\textbf{10: Excellent}\\
\quad • All shapes are drawn correctly and are complete.\\
\quad • Lines are closed where they should be (e.g., in a triangle, square, circle).\\
\quad • The drawing is clear and well-formed.\\[2pt]

\textbf{5--9: Minor to Moderate Flaws}\\
\quad • Shapes are mostly complete but may have small imperfections (e.g., slightly wavy lines, corners not perfectly joined but still closed).\\
\quad • The overall shape is still easily recognizable.\\[2pt]

\textbf{0--4: Severe Drawing Errors}\\
\quad • A primary shape is fundamentally incomplete or malformed.\\
\quad • Examples: a triangle is missing a side, a square has a large gap, a circle is not a closed loop.\\
\quad • The shape is distorted to the point of being unrecognizable.\\[4pt]

\textbf{Output Requirement:}\\
\quad Output \textbf{ONLY} one integer 0--10.\\
  \end{tcolorbox}
  \caption{Prompt for validity check of the rendered original image.}
  \label{tab:checker-image}
\end{table*}


\begin{table*}[t]
  \vspace{-0.3cm}
  \noindent
  \begin{tcolorbox}[                 
    width=\textwidth,        
    before=\noindent,       
    fontupper=\footnotesize, 
    colback=gray!2!white,
    colframe=gray!45!black,
    colbacktitle=gray!15!white,
    coltitle=black,
    fonttitle=\bfseries,
    title={Checker (Visual Tool)}
  ]
You are a STRICT visual tool auditor. You will be given two images: the original problem and the problem with an visual tool drawn. Your task is to compare them and evaluate ONLY the correctness of the visual tool. Output MUST be a single integer score from 0 to 10.\\[4pt]

\textbf{Images:}\\
\quad • Image 1: Original Image\\
\quad • Image 2: Intermediate Visual States\\[4pt]

\textbf{Inputs:}\\
\quad Question: \texttt{"\{question}\}"\\
\quad Original Image Path: \texttt{"\{image\}"}\\
\quad Auxiliary Image Path: \texttt{"\{aux\_image\}"}\\
\quad Auxiliary Line Description: \texttt{"\{visual\_tools\_description\}"}\\
\quad Solution Process that uses the line: \texttt{"\{solution\}"}\\[4pt]

\textbf{Scoring (conservative; compare Image 2 to Image 1):}\\
\quad • Start from 10 ONLY if the line in Image 2 is correctly drawn based on the context from Image 1, matches its description, and is properly used in the solution.\\
\quad • Deduct 4--6 for clear mismatches in type, location, or endpoints (e.g., line from wrong vertex, not tangent).\\
\quad • Deduct 3--5 if a claimed property is not visually/logically satisfied (e.g., line is clearly not perpendicular to the base in Image 1).\\
\quad • Deduct 2--4 if the solution process does not actually reference or use the drawn line as described.\\
\quad • Clamp to [0,10]; prefer lower scores when uncertain.\\[4pt]

\textbf{Output Requirement:}\\
\quad Output \textbf{ONLY} one integer 0--10.\\
  \end{tcolorbox}
  \caption{Prompt for visual tool consistency checking.}
  \label{tab:checker-aux}
\end{table*}
\begin{table*}[t]
  \vspace{-0.5cm}
  \noindent
  \begin{tcolorbox}[                 
    width=\textwidth,        
    before=\noindent,       
    fontupper=\footnotesize, 
    colback=gray!2!white,          
    colframe=gray!45!black,        
    colbacktitle=gray!15!white,    
    coltitle=black,                
    fonttitle=\bfseries,
    title={Repairer}
  ]

You will receive the following inputs: a problem, the corresponding solution steps and answer, the original diagram code, the code for the diagram with visual tools and the image with visual tools. Your task is to use the image, both sets of diagram codes and the context to design or rewrite a new problem, along with its solution and answer, that necessarily depends on the structures or properties introduced by the visual tools. Please follow the steps below to complete the task.\\[4pt]

\textbf{Input includes:}\\
\quad • Problem: \{question\}\\
\quad • Solution steps: \{solution\}\\
\quad • Answer: \{answer\}\\
\quad • Original diagram code: \{code\_ori\}\\
\quad • Visual tools diagram code: \{code\_aux\}\\[4pt]

\textbf{1. Comprehensive Understanding:}\\
\quad • Read and understand the original problem, solution steps, and answer.\\
\quad • Analyze the role and function of the visual tools, using the diagram codes to understand the geometrical structure.\\
\quad • Determine how the visual tools introduce new properties or conclusions that could be leveraged for a novel problem.\\[4pt]

\textbf{2. Create a New Problem:}\\
\quad • Based on the original problem and the visual tools, rewrite or design a new problem that cannot be solved or proven without the visual tools or their structural impact.\\
\quad • The new problem should not be solvable directly from the original diagram code alone, and preferably should not mention the visual tools explicitly.\\
\quad • Ensure that the solution to the new problem crucially uses properties, conclusions, or new geometric configurations stemming from the visual tools.\\
\quad • The new problem may be a proof or computation, but must fundamentally depend on what the visual tools contribute.\\[4pt]

\textbf{3. Instructions for the problem statement:}\\
\quad • Do not include any instructions or hints such as “draw”, “connect”, “extend”, “symmetrize”, “complete”, etc., that refer to constructing visual tools.\\
\quad • If such hints are present, please remove or revise the statement so it gives no guidance on drawing new elements.\\
\quad • The new question should not be answerable by text alone; one should need the accompanying auxiliary-line image.\\[4pt]

\textbf{Final Output Json Format (Output only the final JSON result. No extra explanations or comments):}\\[2pt]
\texttt{\{}\\
\quad \texttt{"id": "\{id\}",}\\
\quad \texttt{"new\_problem": "Newly Designed Problem",}\\
\quad \texttt{"solution\_markdown": "Step-by-step solution in Markdown with LaTeX, where the visual tools are naturally introduced and utilized as part of the reasoning process (CoT)",}\\
\quad \texttt{"answer": "Final Answer"}\\
\texttt{\}}\\

  \end{tcolorbox}
  \caption{Prompt for text–visual consistency repair via question reconstruction.}
  \label{tab:repair}
\end{table*}
\begin{table*}[t]
  \vspace{-0.5cm}
  \noindent
  \begin{tcolorbox}[                 
    width=\textwidth,        
    before=\noindent,       
    fontupper=\tiny, 
    colback=gray!2!white,          
    colframe=gray!45!black,        
    colbacktitle=gray!15!white,    
    coltitle=black,                
    fonttitle=\bfseries,
    title={Progressive Expansion (Parallel Extension)}
  ]
You are an expert exam question designer and visualization engineer. Based on the given original problem, standard answer, auxiliary line image, original drawing code, and auxiliary line drawing code, you need to design either a multiple choice or fill-in-the-blank problem that requires two rounds of auxiliary line code calls to be solved completely. Output the new problem, a clear multi-step auxiliary line drawing plan, the drawing code, and the solution process with the final answer.\\[4pt]

\textbf{Steps:}\\
\quad • You need to read and understand all inputs. Use the provided auxiliary line image to interpret the problem, the purpose of the auxiliary line, and the solution process.\\
\quad • The existing auxiliary line provides one key conclusion for problem-solving; another key conclusion must come from the newly introduced auxiliary line. The new problem you create must depend on at least these two key conclusions to be solved.\\
\quad • Read the auxiliary line image code; based on this image, write executable Python code to draw the new auxiliary lines. You may refer to the original drawing code to help interpret the geometric structure and coordinates of the points and lines in the original and auxiliary images.\\
\quad • Ensure the new problem maintains the original point naming. Do not pose questions about nonexistent geometric elements.\\
\quad • The answer should not be directly obtainable from the original or auxiliary diagram, and each round of auxiliary lines must be necessary and enhance subsequent reasoning.\\
\quad • Generate the output: problem statement, reasoning for why multiple rounds of auxiliary lines are necessary, drawing code, detailed problem-solving process and final answer.\\[4pt]

\textbf{Rules for introducing new visual tools:}\\
\quad • The new auxiliary line must be independent from the original auxiliary line; both contribute separate key conclusions that are jointly necessary for the solution.\\
\quad • Do not design multiple sub-problems; each step must serve the solution to the same final problem.\\
\quad • The new auxiliary line can leverage the structure of the original auxiliary line — it can be a simple connection, or a new point constructed using the original auxiliary line (e.g. midpoint, projection, external point), a new segment/intersection, or projection.\\
\quad • Load auxiliary line image and draw through the code below:\\
\qquad \texttt{image\_path = \{aux1\}}\\
\qquad \texttt{img = plt.imread(image\_path)}\\
\qquad \texttt{h, w = img.shape[:2]}\\
\qquad \texttt{fig = plt.figure()}\\
\qquad \texttt{ax = fig.add\_axes([0,0,1,1])}\\
\qquad \texttt{ax.set\_xlim(0, w)}\\
\qquad \texttt{ax.set\_ylim(h, 0)}\\
\qquad \texttt{ax.axis('off')}\\
\qquad \texttt{ax.set\_aspect('equal')}\\
\qquad \texttt{ax.imshow(img)}\\
\quad • End the code with \texttt{plt.show()}.\\[4pt]

\textbf{Instructions for the problem statement:}\\
\quad • The new problem must be based on the original diagram, but \textbf{must not reference the original problem, auxiliary line, or hint at the construction} (e.g. do not provide the original auxiliary line or answer as given conditions). The original auxiliary line and answer must be included as steps necessary for solving the new problem.\\
\quad • All variable names used in the new problem statement, solution, and auxiliary line drawing code must strictly follow the naming conventions established in the original diagram and the provided drawing code.\\
\quad • The problem statement \textbf{must not include any explicit or implicit hints} for construction or drawing, such as ``draw'', ``connect'', ``extend'', ``symmetry'', ``complete'', etc.\\[4pt]

\textbf{Available inputs:}\\
\quad • Original problem text: ``\{question\}''\\
\quad • Original solution path: ``\{solution\}''\\
\quad • Original standard answer: ``\{answer\}''\\
\quad • Original drawing code: ``\{code\_ori\}''\\
\quad • Original auxiliary tools description: ``\{auxiliary\_tools\_description\}''\\
\quad • Auxiliary line code: ``\{code\_aux\}''\\
\quad • Auxiliary line image\\[4pt]

\textbf{Final Output Json Format (Output only the final JSON result. No extra explanations or comments):}\\[2pt]
\texttt{\{}\\
\quad \texttt{"id": "\{id\}",}\\
\quad \texttt{"new\_problem": "Newly Designed Problem",}\\
\quad \texttt{"problem\_type": "multiple choice or fill\_in\_blank",}\\
\quad \texttt{"new\_auxiliary\_tools\_description": "Brief description of the new auxiliary tools to be added (what, where and functions)",}\\
\quad \texttt{"python\_code\_auxiliary": "Pure Python code as a string, directly runnable (loads the saved figure image via plt.imread and overlays auxiliary tools in pixel coordinates)",}\\
\quad \texttt{"solution\_markdown": "Step-by-step solution in Markdown with LaTeX, where the auxiliary lines are naturally introduced and utilized as part of the reasoning process (CoT)",}\\
\quad \texttt{"final\_answer": "Final Answer"}\\
\texttt{\}}\\
  \end{tcolorbox}
  \caption{Prompts for parallel extension.}
  \label{tab:parallel-extension-prompt}
\end{table*}

\begin{table*}[t]
  \vspace{-0.5cm}
  \noindent
  \begin{tcolorbox}[                 
    width=\textwidth,        
    before=\noindent,       
    fontupper=\tiny, 
    colback=gray!2!white,          
    colframe=gray!45!black,        
    colbacktitle=gray!15!white,    
    coltitle=black,                
    fonttitle=\bfseries,
    title={Progressive Expansion (Sequential Extension)}
  ]
You are an expert exam question designer and visualization engineer. Based on the given original problem, standard answer, auxiliary line image, original drawing code, and auxiliary line drawing code, you need to design either a multiple choice or fill-in-the-blank problem that requires two rounds of auxiliary line code calls to be solved completely. Output the new problem, a clear multi-step auxiliary line drawing plan, the drawing code, and the solution process with the final answer.\\[4pt]

\textbf{Steps:}\\
\quad • You need to read and understand all inputs. Use the provided auxiliary line image to interpret the problem, the purpose of the auxiliary line, and the solution process.\\
\quad • The existing auxiliary line is one key conclusion for solving; the other key conclusion must rely on a newly introduced auxiliary line. The new question you provide must require at least two key conclusions or constructions to be solved.\\
\quad • Read the auxiliary line image and, based on it, write executable Python code to draw a new auxiliary line. You may use the original code to help understand the points, lines, coordinates, and geometric constructions in the original and auxiliary images.\\
\quad • Ensure the new problem maintains the original point naming. Do not pose questions about nonexistent geometric elements.\\
\quad • The answer should not be directly obtainable from the original or auxiliary diagram, and each round of auxiliary lines must be necessary and enhance subsequent reasoning.\\
\quad • Generate the output: problem statement, reasoning for why multiple rounds of auxiliary lines are necessary, drawing code, detailed problem-solving process and final answer.\\[4pt]

\textbf{Rules for introducing new visual tools:}\\
\quad • The new auxiliary line must be closely related to the original auxiliary line or the original problem's key conclusion. It must be constructed based on knowledge obtained from the original auxiliary line or key conclusion. For example, if the original auxiliary line is \texttt{DM}, a new auxiliary line could be the perpendicular to \texttt{DM}.\\
\quad • Do not design multiple small subproblems; every step must serve the solution to a single, unified problem.\\
\quad • The new auxiliary line can leverage the structure of the original auxiliary line — it can be a simple connection, or a new point constructed using the original auxiliary line (e.g. midpoint, projection, external point), a new segment/intersection, or projection.\\
\quad • Load auxiliary line image and draw through the code below:\\
\qquad \texttt{image\_path = \{aux1\}}\\
\qquad \texttt{img = plt.imread(image\_path)}\\
\qquad \texttt{h, w = img.shape[:2]}\\
\qquad \texttt{fig = plt.figure()}\\
\qquad \texttt{ax = fig.add\_axes([0,0,1,1])}\\
\qquad \texttt{ax.set\_xlim(0, w)}\\
\qquad \texttt{ax.set\_ylim(h, 0)}\\
\qquad \texttt{ax.axis('off')}\\
\qquad \texttt{ax.set\_aspect('equal')}\\
\qquad \texttt{ax.imshow(img)}\\
\quad • End the code with \texttt{plt.show()}.\\[4pt]

\textbf{Instructions for the problem statement:}\\
\quad • The new problem must be based on the original diagram, but \textbf{must not reference the original problem, auxiliary line, or hint at the construction} (e.g. do not provide the original auxiliary line or answer as given conditions). The original auxiliary line and answer must be included as steps necessary for solving the new problem.\\
\quad • All variable names used in the new problem statement, solution, and auxiliary line drawing code must strictly follow the naming conventions established in the original diagram and the provided drawing code.\\
\quad • The problem statement \textbf{must not include any explicit or implicit hints} for construction or drawing, such as ``draw'', ``connect'', ``extend'', ``symmetry'', ``complete'', etc.\\[4pt]

\textbf{Available inputs:}\\
\quad • Original problem text: ``\{question\}''\\
\quad • Original solution path: ``\{solution\}''\\
\quad • Original standard answer: ``\{answer\}''\\
\quad • Original drawing code: ``\{code\_ori\}''\\
\quad • Original auxiliary tools description: ``\{auxiliary\_tools\_description\}''\\
\quad • Auxiliary line code: ``\{code\_aux\}''\\
\quad • Auxiliary line image\\[4pt]

\textbf{Final Output Json Format (Output only the final JSON result. No extra explanations or comments):}\\[2pt]
\texttt{\{}\\
\quad \texttt{"id": "\{id\}",}\\
\quad \texttt{"new\_problem": "Newly Designed Problem",}\\
\quad \texttt{"problem\_type": "multiple choice or fill\_in\_blank",}\\
\quad \texttt{"new\_auxiliary\_tools\_description": "Brief description of the new auxiliary tools to be added (what, where and functions)",}\\
\quad \texttt{"python\_code\_auxiliary": "Pure Python code as a string, directly runnable (loads the saved figure image via plt.imread and overlays auxiliary tools in pixel coordinates)",}\\
\quad \texttt{"solution\_markdown": "Step-by-step solution in Markdown with LaTeX, where the auxiliary lines are naturally introduced and utilized as part of the reasoning process (CoT)",}\\
\quad \texttt{"final\_answer": "Final Answer"}\\
\texttt{\}}\\
  \end{tcolorbox}
  \caption{Prompts for sequential extension.}
  \label{tab:sequential-extension-prompt}
\end{table*}
\begin{table*}[t]
  \vspace{-0.5cm}
  \noindent
  \begin{tcolorbox}[                 
    width=\textwidth,        
    before=\noindent,       
    fontupper=\footnotesize, 
    colback=gray!2!white,
    colframe=gray!45!black,
    colbacktitle=gray!15!white,
    coltitle=black,
    fonttitle=\bfseries,
    title={Perception Data Synthesis (Elements \& Visual tag)}
  ]
You are an expert in constructing visual representations across diverse domains. Based on the specified knowledge point, number of elements, and relationships between elements, please select elements that are consistent with the knowledge point to generate the figure. Output executable Python drawing code and precise data annotations.\\[3pt]

\textbf{1. Element Generation Rules}\\
\quad • The basic figure must completely reflect the specified knowledge point.\\
\quad • For special knowledge points (tables, stem--leaf plots, mazes, \dots), use dedicated drawing tools (e.g., \texttt{ax.table}) rather than composing basic elements manually.\\
\quad • For statistical tables: the number of elements should equal the number of table cells plus required text/labels. No extra points/lines.\\
\quad • All basic elements must be clearly annotated in the element property list.\\
\quad • A line may be a diagonal, or a segment with existing endpoints.\\
\quad • Angles and extra lines must be derived from existing vertices.\\
\quad • Added complexity must not obscure the assessment of the knowledge point. Completely random/unrelated elements are forbidden.\\
\quad • No element may exceed the canvas boundary.\\[3pt]

\textbf{2. Element Definition \& Attribute Rules}\\
\quad • \textbf{Point:} \texttt{{"type":"point","x":...,"y":...,"label":"A","semantic":""}}\\
\quad • \textbf{Line:} \texttt{{"type":"line","x1":...,"y1":...,"x2":...,"y2":...,"semantic":""}}\\
\quad • \textbf{Angle:} \texttt{{"type":"angle","vertex":[x,y],"dir1":[dx1,dy1],"dir2":[dx2,dy2]}}\\
\quad • \textbf{Circle:} \texttt{{"type":"circle","cx":...,"cy":...,"r":...}}\\
\quad • \textbf{Function Curve:} \texttt{{"type":"function\_curve","expression":"y=sin(x)","domain":[...],"style":"solid"}}\\
\quad • \textbf{Text:} \texttt{{"type":"text","content":"...", "position":[x,y]}}\\
\quad • \textbf{Symbol:} \texttt{{"type":"symbol","position":[x,y],"symbol\_type":"perpendicular"}}\\[2pt]
\quad All elements must include a \texttt{semantic} field (write “None” if unnecessary).\\[3pt]

\textbf{3. Coordinate \& Size Requirements}\\
\quad • All coordinates must be absolute pixel values (integers).\\
\quad • No out-of-bound elements; circles must stay within canvas.\\
\quad • Angle arcs must be computed from the given direction vectors.\\[3pt]

\textbf{4. Drawing \& Coding Specifications}\\
\quad • Must output fully executable Python code using matplotlib.\\
\quad • Required boilerplate:  \\
\quad\quad \texttt{w, h =}\\
\quad\quad \texttt{DPI =}\\
\quad\quad \texttt{fig = plt.figure(figsize=(w/DPI, h/DPI), dpi=DPI)}\\
\quad\quad \texttt{ax = fig.add\_axes([0,0,1,1])}\\
\quad\quad \texttt{ax.set\_xlim(0, w)}\\
\quad\quad \texttt{ax.set\_ylim(h, 0)}\\
\quad\quad \texttt{ax.axis('off')}\\
\quad\quad \texttt{ax.set\_aspect('equal')}\\
\quad • Do NOT use \texttt{plt.tight\_layout()}.\\
\quad • End with \texttt{plt.show()}.\\
\quad • The diagram must not contain extra explanatory text.\\[4pt]

\textbf{5. Output Format}\\
Output must follow this JSON format (no extra text):\\[2pt]
\texttt{\{}\\
\quad \texttt{"id": "{new\_id}",}\\
\quad \texttt{"knowledge\_point": "{knowledge\_point}",}\\
\quad \texttt{"element\_num": "{element\_num}",}\\
\quad \texttt{"python\_code": "...",}\\
\quad \texttt{"element\_arrtibute": "..."}\\
\texttt{\}}\\[3pt]

\textbf{Input Arguments:}\\
\quad • Knowledge Concept: \texttt{"{knowledge\_concept}"}\\
\quad • Number of elements: \texttt{"greater than \{element\_num\}"}\\
\quad • Element relationships: \texttt{"{relation}"}\\
  \end{tcolorbox}
  \caption{Prompt for element and perception tag construction.}
  \label{tab:figure-construction-prompt}
\end{table*}

\begin{table*}[t]
  \vspace{-0.5cm}
  \noindent
  \begin{tcolorbox}[                 
    width=\textwidth,        
    before=\noindent,       
    fontupper=\footnotesize, 
    colback=gray!2!white,
    colframe=gray!45!black,
    colbacktitle=gray!15!white,
    coltitle=black,
    fonttitle=\bfseries,
    title={Perception Data Synthesis (Surface-level)}
  ]
You are a point-level visual QA data generation expert. Based on the image description and the names of the points/elements to be detected, please generate question–answer data. The question should ask for the locations of these points or vertices in the image, and the answer should provide their coordinates in dictionary format.\\[4pt]

\textbf{Input:}\\
The element annotation information in the image (absolute coordinates) is:\\
\quad \texttt{\{element\_arrtibute\}}\\[4pt]

\textbf{Example:}\\
\quad Question: Detect and output the pixel coordinates of points A, B, and C in the image (dictionary format is recommended, e.g., \texttt{\{\,'A': (x1, y1), 'B': (x2, y2), 'C': (x3, y3)\,\}}).\\
\quad Answer: \texttt{\{\,'A': (x1, y1), 'B': (x2, y2), 'C': (x3, y3)\,\}}\\[4pt]

\textbf{Specification notes:}\\
\quad • If there are no explicit geometric shapes, ask perception-related questions, such as the coordinates of the points where a function achieves a specific value in the image, or comparing/calculating statistical data shown in image tables.\\
\quad • Always provide the answers as absolute coordinates (pixel positions) of relevant targets or data within the image, rather than abstract mathematical results.\\[4pt]

\textbf{Output Format:}\\
Final output must follow this JSON format (output only the JSON, no extra explanations or comments):\\[2pt]
\texttt{\{}\\
\quad \texttt{"id": "\{id\}",}\\
\quad \texttt{"question": "",}\\
\quad \texttt{"answer": "",}\\
\quad \texttt{"type": "Surface Level Perception QA"}\\
\texttt{\}}\\

  \end{tcolorbox}
  \caption{Prompt for surface-level perception Q\&A construction.}
  \label{tab:surfaceqa}
\end{table*}

\begin{table*}[t]
  \vspace{-0.5cm}
  \noindent
  \begin{tcolorbox}[                 
    width=\textwidth,        
    before=\noindent,       
    fontupper=\footnotesize, 
    colback=gray!2!white,
    colframe=gray!45!black,
    colbacktitle=gray!15!white,
    coltitle=black,
    fonttitle=\bfseries,
    title={Perception Data Synthesis (Semantic-level)}
  ]
You are a semantic-level visual QA data generation expert. Please create QA pairs for specified spatial–semantic points based on the structural semantic annotations of the input image. The question should ask for detection of specific structural feature points in the image (e.g., ``the top-left vertex of the cube''), and the answer should return their coordinates.\\[4pt]

\textbf{Input:}\\
The element annotation information in the image (absolute coordinates) is:\\
\quad \texttt{\{element\_arrtibute\}}\\[4pt]

\textbf{Example:}\\
\quad Question: Detect and output the pixel coordinates of the top-left front corner of the cube in the image (dictionary format is recommended, e.g., \texttt{\{\,'Top-left vertex': (x, y)\,\}}).\\
\quad Answer: \texttt{\{\,'Top-left vertex': (x, y)\,\}}\\[4pt]

\textbf{Specification notes:}\\
\quad • Use spatial descriptions or mathematical definitions (e.g., ``top-left front corner of the upper surface'', ``vertex of a triangle'') in the \textbf{question}.\\
\quad • If the element annotation information provides explicit naming or numbering for the corresponding vertex (e.g., \texttt{A}, \texttt{B}, \texttt{P1}, \texttt{V3}), use these symbols as the standard expression in the \textbf{answer}. Do \textbf{not} use these explicit names in the question.\\
\quad • If there are no explicit geometric shapes, ask perception-related questions, such as the coordinates of the points where a function achieves a specific value in the image, or comparing/calculating statistical data shown in image tables.\\
\quad • Always provide the answers as absolute coordinates (pixel positions) of relevant targets or data within the image, rather than abstract mathematical results.\\[4pt]

\textbf{Output Format:}\\
Final output must follow this JSON format (output only the JSON, no extra explanations or comments):\\[2pt]
\texttt{\{}\\
\quad \texttt{"id": "\{id\}",}\\
\quad \texttt{"question": "",}\\
\quad \texttt{"answer": "",}\\
\quad \texttt{"type": "Semantic Level Reasoning QA"}\\
\texttt{\}}\\

  \end{tcolorbox}
  \caption{Prompt for semantic-level visual Q\&A construction.}
  \label{tab:semanticqa}
\end{table*}

\begin{table*}[t]
  \vspace{-0.5cm}
  \noindent
  \begin{tcolorbox}[                 
    width=\textwidth,        
    before=\noindent,       
    fontupper=\footnotesize, 
    colback=gray!2!white,
    colframe=gray!45!black,
    colbacktitle=gray!15!white,
    coltitle=black,
    fonttitle=\bfseries,
    title={Perception Data Synthesis (Integrated reasoning)}
  ]
You are an expert in generating QA data for visual perception and computation tasks. Please combine the input image's structural information and geometric reasoning to generate visual QA pairs such as finding the pixel coordinates of the center point of a square shown in the image.\\[4pt]

\textbf{Input:}\\
The element annotation information in the image (absolute coordinates) is:\\
\quad \texttt{\{element\_arrtibute\}}\\[4pt]

\textbf{Example:}\\
\quad Question: Detect and output the pixel coordinates of the center point of the square in the image (dictionary format recommended, e.g., \texttt{\{\,'Center': (x, y)\,\}}).\\
\quad Answer: \texttt{\{\,'Center': (x, y)\,\}}\\[4pt]

\textbf{Specification notes:}\\
\quad • Use spatial descriptions or mathematical definitions (e.g., ``foot of the perpendicular from vertex A of the trapezoid to base CD'', ``incenter of the triangle'') in the \textbf{question}.\\
\quad • If the element annotation information provides explicit naming or numbering for corresponding points (e.g., \texttt{A}, \texttt{B}, \texttt{O}, \texttt{V3}), use these annotation symbols in the \textbf{answer}, but \textbf{never} in the question.\\
\quad • If the image contains no explicit geometric shapes, ask perception-related tasks such as:  \\
\qquad – locating where a function achieves a specific value;\\
\qquad – detecting coordinates corresponding to statistical data in charts or tables.\\
\quad • All answers must be absolute pixel coordinates of the target elements, not abstract math results.\\[4pt]

\textbf{Output Format:}\\
The final output must follow this JSON format (output only the JSON, no extra explanations or comments):\\[2pt]
\texttt{\{}\\
\quad \texttt{"id": "\{id\}",}\\
\quad \texttt{"question": "",}\\
\quad \texttt{"answer": "",}\\
\quad \texttt{"type": "Integrated Reasoning QA"}\\
\texttt{\}}\\

  \end{tcolorbox}
  \caption{Prompt for integrated reasoning Q\&A construction.}
  \label{tab:Integratedqa}
\end{table*}

\clearpage
\clearpage


\end{document}